\documentclass[preprint,3p,review]{elsarticle}

\usepackage{ecrc}

\volume{00}

\firstpage{1}

\runauth{G. Norman et al.}

\usepackage{amssymb}

\usepackage{amsmath}
\usepackage{algorithm}
\usepackage{algorithmic}
\usepackage{caption}
\usepackage{xcolor}

\usepackage{lineno}

\usepackage[figuresright]{rotating}

\newcommand{\du}{\ensuremath{\mathrm{d}}}
\newcommand{\Nc}{\ensuremath{\mathcal{N}}}
\newcommand{\Np}{\ensuremath{\mathcal{N}}^{\phi}}
\newcommand{\ut}{\ensuremath{u^{\theta}}}
\newcommand{\up}{\ensuremath{u^{\phi}}}
\newcommand{\fj}{\ensuremath{r_{j}^{\theta, \phi}}}
\newcommand{\lamo}{\ensuremath{\lambda^{0}}}
\newcommand{\val}{\ensuremath{L_{\text{val}}^{n_x}(\phi)}}
\newcommand{\train}{\ensuremath{L_{\text{train}}(\theta, \phi)}}
\newcommand{\rell}{\ensuremath{\ell^{2}_{\mathrm{rel}}}}
\newcommand{\utrain}{\ensuremath{u_{0}^{\mathrm{train}}}}
\newcommand{\utest}{\ensuremath{u_{0}^{\mathrm{test}}}}
\DeclareMathOperator*{\argminop}{arg\,min}

\newcommand{\norm}[1]{\left\lVert#1\right\rVert}
\newcommand{\Nf}{\ensuremath{N_r}}

\begin{document}

\begin{frontmatter}



\dochead{}

\title{Constrained or Unconstrained? Neural-Network-Based Equation Discovery from Data}



\author[cu]{Grant Norman}
\author[sandia]{Jacqueline Wentz}
\author[sandia]{Hemanth Kolla}
\author[cu]{Kurt Maute}
\author[cu]{Alireza Doostan\corref{cor5}}%
\cortext[cor5]{Corresponding author}

\address[cu]{Ann and H.J. Smead Aerospace Engineering Sciences, University of Colorado Boulder, CO 80303}
\address[sandia]{Sandia National Laboratories, Livermore, CA 94551}

\begin{abstract}
Throughout many fields, practitioners often rely on differential equations to model systems.
Yet, for many applications, the theoretical derivation of such equations and/or accurate resolution of their solutions may be intractable.
Instead, recently developed methods, including those based on parameter estimation, operator subset selection, and neural networks, allow for the data-driven discovery of both ordinary and partial differential equations (PDEs), on a spectrum of interpretability.
The success of these strategies is often contingent upon the correct identification of representative equations from noisy observations of state variables and, as importantly and intertwined with that, the mathematical strategies utilized to enforce those equations. 
Specifically, the latter has been commonly addressed via unconstrained optimization strategies.
Representing the PDE as a neural network, we propose to discover the PDE by solving a constrained optimization problem and using an intermediate state representation similar to a Physics-Informed Neural Network (PINN).
The objective function of this constrained optimization problem promotes matching the data, while the constraints require that the PDE is satisfied at a number of spatial collocation points.
We present a penalty method and a widely used trust-region barrier method to solve this constrained optimization problem, and we compare these methods on numerical examples.
Our results on the Burgers' and the Korteweg-De Vreis equations demonstrate that the latter constrained method outperforms the penalty method, particularly for higher noise levels or fewer collocation points.
This work motivates further exploration into using sophisticated constrained optimization methods in scientific machine learning, as opposed to their commonly used, penalty-method or unconstrained counterparts.
For both of these methods, we solve these discovered neural network PDEs with classical methods, such as finite difference methods, as opposed to PINNs-type methods relying on automatic differentiation.
We briefly highlight how simultaneously fitting the data while discovering the PDE improves the robustness to noise and other small, yet crucial, implementation details.
\end{abstract}


\end{frontmatter}


\section{Introduction}
\label{sec:intro}

Differential equations are used ubiquitously throughout science and engineering.
For instance, continuum mechanics, finance, and quantum mechanics rely upon ordinary and partial differential equations (ODEs and PDEs).
Scientists, engineers, and mathematicians use these differential equations to gain deeper insights into a particular system, predict the behavior of existing physical systems later in time, or simulate hypothetical configurations.
Stated differently, differential equations are already a valuable tool in countless applications.
While most of the aforementioned fields theoretically derive these equations, there are still many situations where the governing equations are unknown.
Given the success of using differential equations in these fields, it is reasonable to hope for some problems to be described by differential equations that have yet to be discovered.
Rather than relying on potentially intractable theoretical derivations, recent advances at the interface of machine learning, science, and engineering have proposed learning ODEs and PDEs from data; see, e.g., \cite{brunton_discovering_2016, chen_neural_2019, rackauckas_universal_2021, raissi_hidden_2020, raissi_deep_2018, wu_data-driven_2020}.
Recently, this notion has become increasingly attractive due to the increasing availability of data, the growth of open-source machine learning tools, and improvements in the hardware used for machine learning.
As opposed to traditional machine learning, differential equations allow for further interpretation and generalization, while usually requiring less data and smaller models.
For instance, by changing the geometry, initial conditions, or boundary conditions, the same differential equations can model multiple systems governed by the same phenomena.
In some applications, a differential equation may be used to simplify a more computationally expensive model.
This differential equation can then be used, for instance, as a surrogate model enabling multi-fidelity uncertainty quantification or optimization of the original expensive model.
In summary, learning a differential equation is relevant to many areas of science and engineering, which we will focus on in the present study.

A number of challenges still exist for algorithmically learning PDEs from data.
First, the available data is often noisy, and the computation of derivatives from noisy data can lead to poor estimates that in turn lead to inaccurate equations \cite{cortiella_priori_2022}.
Second, the functional form of the PDE must often be assumed, up to a number of parameters, thus imposing a potentially unjustified prior.
For instance, some methods require the construction of a ``library'' representing different candidate terms in the PDE \cite{rudy_deep_2019, rudy_data-driven_2016, messenger_weak_2021, wentz_derivative-based_2023-1}.
These methods often involve a further sparsity assumption, leading to the discovery of interpretable PDEs.
Third, most of the existing methods rely on penalty formulations or compound loss functions to weakly encourage adherence to the governing equations.

To clarify these points, suppose we seek a general PDE of the form
\begin{equation}
    u_t = \Nc (x, t, u, u_x, u_{xx}, \dots) ,
    \label{eq:intro:main_pde}
\end{equation}
governing the state given by scalar $u(x,t)$.
Note that in some cases, we may need more restrictive requirements on the solution space in order to write the PDE in this form, for instance assuming $u \in \mathcal{C}^2$ as opposed to $u \in \mathcal{H}^2$ \cite{messenger_weak_2021}, but this distinction is not pertinent to our discussion.
In general, $\Nc$ can be parameterized through a variety of methods, but first, the inputs to $\Nc$ must be decided.
Some prior knowledge should be used to determine if the time evolution depends explicitly on $x$ or $t$, and how many spatial derivatives to include.
Then, $\Nc$ can be parameterized, for instance using a symbolic expression \cite{schaeffer_learning_2017}, a polynomial expansion \cite{rudy_data-driven_2016},
a symbolic neural network \cite{long_pde-net_2019},
or a rational neural network \cite{stephany_pde-learn_2023}.
In this work, inspired by the deep hidden physics approach of \cite{raissi_deep_2018}, we focus on a (standard) neural network functional form, denoting this expression as $\Np$, where $\phi$ are the parameters of the representing $\Nc$.
Unlike the aforementioned methods, we are \textit{not} concerned with the interpretability of our PDE in this method.
However, this more general representation allows for a more diverse class of functions to be represented, as opposed to just polynomials up to some total degree, and a non-interpretable PDE can still be used to forecast and make new predictions.
We emphasize that for many \textit{other} applications, the PDE may be partially known, or more information regarding the system may be available.
This information should be encoded into the parameterization of $\Np$ \cite{rackauckas_universal_2021}.
We concern ourselves with the case where very little is known about the PDE, and thus very little information can be encoded in the parameterization of $\Np$.
Applications in biology and finance may fall into this category.

Next, to overcome the difficulty of working with noisy data, while there are many possible \textit{a priori} denoising methods \cite{cortiella_priori_2022}, we elect to approximate the data with a smooth function, and use exact derivatives of this function to estimate derivatives of the true state, \textit{along with learning the PDE}.
Keeping in theme with our choice of $\Np$, we use a neural network $\ut$ to approximate the data and apply automatic differentiation to compute its derivatives.
As opposed to \cite{raissi_deep_2018}, we require $\ut$ to satisfy the governing PDE, through a constrained formulation, similar to \cite{hokanson_simultaneous_2022}, which proposed using a preconditioned sequential quadratic programming approach to discover governing ODEs.
For our application of PDE discovery, we compare a loosely-constrained penalty method against a trust-region barrier method.

This work tackles a different problem than that of Physics-Informed Neural Networks (PINNs) \cite{raissi_physics-informed_2019}, although there are some similarities between the methods.
PINNs represent the state of a system governed by a PDE as a neural network $\ut$, which we also do.
Both methods compute residuals of a PDE at a number of collocation points throughout the spatial-temporal domain.
Then, the boundary condition and initial conditions are sampled and compared to their prediction through $\ut$.
In PINNs, the PDE residuals, initial and boundary conditions, and (potentially) other measurements of the state are combined through a compound loss function.
The parameters $\theta$ are then trained through some iterative, gradient-based method, such as Adam \cite{kingma_adam_2014}.
There have been many improvements and extensions proposed for PINNs beyond this basic outline \cite{cuomo_scientific_2022}.
Most relevant to our work, Self-adaptive PINNs \cite{Mcclenny2022} posed training PINNs as a constrained optimization problem.
Motivated by the success of attention in other fields \cite{vaswani_attention_2023},
the authors proposed solving this constrained optimization problem with a penalty method (see Sec. \ref{sec:sa}).
Subsequent works have proposed using Lagrangian methods instead \cite{Basir2022_PECANNs, lu_physics-informed_2021}, and satisfying certain initial and boundary conditions by construction of the state \cite{Wang2022_respecting, lu_physics-informed_2021}.
Along these lines, while not a PINN, \cite{xu_physics_2022} proposed an inverse-problem formulation where the state is constrained to satisfy the PDE.
Specifically for hyperbolic PDEs, others have proposed different conservative approaches \cite{jagtap_conservative_2020, hansen_learning_2023}.

But, for (forward) PINNs, the governing PDE is entirely known, which differs from our case.
The objective of PINNs is to construct $\ut$ as the solution to the known PDE,
while we merely use $\ut$ as an intermediary to \textit{discover} the PDE.
Thus, as opposed to PINNs, our inverse problem \textit{requires} samples of the state on the interior of the domain, in addition to boundary and initial condition information.
For PINNs, interior data can be incorporated, but for our PDE discovery problem, such interior data is absolutely essential.
Further, unlike finding the unique solution to a well-posed PDE, this problem may be fundamentally ill-posed, having infinitely many solutions (even with noiseless interior data) \cite{Aster_rank_def_ill_cond}.
That is, there may be infinitely many PDEs that have solutions matching the data, the initial conditions, and the boundary conditions, both from a structural and practical identifiability perspective \cite{raue_structural_2009, miao_identifiability_2011}.

Further, our problem is different from many parameter estimation problems, as the entire form of the PDE is unknown \cite{vogel_parameter_1979}.
This problem may be seen as a specific case of operator learning \cite{lu_deeponet_2021, li_fourier_2021}, where we seek to approximate the operator mapping from initial conditions and boundary conditions to the solution.
Yet, we impose a certain structure on this operator through the learned PDE.
Flow map learning \cite{churchill_flow_2023, wu_data-driven_2020, chen_deep_2022} uses a similar approach, where the form of the PDE is not explicitly represented, but where the action of the time-evolution operator is approximated.
As opposed to operator learning, our explicit representation of the PDE allows us to train our ``operator'' on just one set of initial and boundary conditions, although with a greater cost at evaluation time due to numerically solving the learned PDE $\Np$.
Ideally, by learning the form of the PDE (as opposed to an operator or flow map) we can more seamlessly change initial and boundary conditions, modify the simulation domains, or coarsen and refine meshes.
Further, this approach allows for more interpretation, although it is not as readily interpretable as SINDy-like methods \cite{rudy_data-driven_2016, messenger_weak_2021}.

With these connections aside, this work is concerned with how to use the data to determine $\Np$ from noisy data, while using $\ut$ as an intermediate, denoised representation of the state.
We present two methods for gradient-based training: a loosely-constrained penalty method and a trust-region, barrier method.
We also review a classical numerical method, which we use to solve the neural network PDE $\Np$.
We touch on the new challenges that arise, overview how the choice of the validation loss can mitigate these issues, and discuss other implementation details.
We find that by constraining the PDE residuals with a trust-region barrier method, the discovery is significantly more robust to noise.

The rest of the paper is organized as follows.
In Sec. \ref{sec:background}, we provide the necessary background on the method of lines and neural-network-based equation discovery.
Following this, in Sec. \ref{sec:methods}, we present the general constrained formulation of the problem before posing a penalty-method approach and a barrier-method approach.
We also overview a few critical implementation details, such as simultaneously training $\Np$ and $\ut$.
Sec. \ref{sec:results} then introduces two example problems and compares these two approaches on them.
We conclude by comparing the performance of the two methods, summarizing other major takeaways, and presenting future research directions and potential improvements.

\section{Background}
\label{sec:background}
In this section, we make a brief aside to introduce the method of lines and then present a baseline method for learning neural network PDEs, first proposed in \cite{raissi_deep_2018}.

\subsection{Method of Lines}
\label{sec:background:mol}

The method of lines discretizes a PDE over the spatial dimension(s) to yield a system of ODEs and then integrates this system in time \cite{leveque_advection_2007}.
Consider a PDE of the form given by Eq. (\ref{eq:intro:main_pde}), with initial condition $u(x,0) = u^0(x)$, an equi-spaced 1D spatial mesh $\{ x_k \}_{k=0}^{n_x}$, and zero Dirichlet boundary conditions at $x_0$ and $x_{n_x}$.
Through the method of lines, we apply spatial finite differences to yield a system of ODEs.
Denoting ${v}_k = u(x_k, \cdot )$ and the uniform mesh spacing as $\Delta x$, the standard 2nd-order, centered finite differences give
\begin{align*}
    u_x (x_k, \cdot) & \approx \frac{{v}_{k+1} - {v}_{k-1}}{2 \Delta x}, \quad k = 1, ..., n_x - 1,\\
    u_{xx} (x_k, \cdot) & \approx \frac{{v}_{k+1} - 2 {v}_k + {v}_{k-1}}{( \Delta x ) ^ 2}, \quad k = 1, ..., n_x - 1 .
\end{align*}
Using these approximations, we arrive at a system of ODEs, where the evolution of each state is given by
\begin{equation}
    \frac{\du {{v}}_k}{\du t} = \Nc \left (
    x_k,
    t, 
    {v}_k,
    \frac{{v}_{k+1} - {v}_{k-1}}{2 \Delta x},
     \frac{{v}_{k+1} - 2 {v}_k + {v}_{k-1}}{( \Delta x ) ^ 2}
    \right ),
    \quad k = 1, ..., n_x - 1 ,
    \label{eq:method_of_lines}
\end{equation}
where ${v}_0 = {v}_{n_x} = 0$ due to the Dirichlet boundary conditions.
The initial condition of $v_k$ is given by evaluating $u^0 (x_k)$ for $k = 1, ..., n_x - 1$.
Thus, we can now integrate these equations in time to obtain a discretized solution to the original PDE.
Without loss of generality, this method can be applied to PDEs with higher order spatial derivatives and different boundary conditions.
In summary, the method of lines is a simple discretization in space which allows us to represent our state discretely as a system of coupled ODEs, with accompanying initial conditions.

\subsection{Deep Hidden Physics Models}
\label{sec:dhpm}
Here, we review Deep Hidden Physics Models \cite{raissi_deep_2018}, before continuing to our extensions.
This work establishes the fundamentals of neural-network-based equation discovery.
Suppose we are given (potentially) noisy measurements at various locations and times of some state $u(x,t)$, which has spatio-temporal domain $\Omega \times [0,T]$.
Such a collection forms our potentially unstructured dataset $\{ (x_i, t_i), u_i \}_{i=1}^{N_u}$, where $N_u$ is the total number of points.
Next, we create a neural network representation of the state, denoted by $\ut (x,t)$, which has parameters $\theta$.
One possible approach is to first fit $\ut$ to the data, iteratively minimizing the standard mean squared error (MSE) loss function
\begin{equation}
    L_\text{MSE} (\theta) = \frac{1}{N_u} \sum_{i=1}^{N_u} \left ( u_i - \ut (x_i, t_i) \right )^2 .
    \label{eq:mse}
\end{equation}

Central to this method, we assume that $u(x,t)$ evolves according to a PDE, such that Eq. (\ref{eq:intro:main_pde}) holds.
We also assume which factors influence the time evolution of $u$, and then parameterize the right-hand side as a neural network with parameters $\phi$.
For example, assuming that $\Nc$ uses up to two spatial derivatives and does not explicitly depend on $x$ or $t$, we have that $u_t = \Np (u, u_x, u_{xx})$.
The decision of which inputs to include in $\Np$ is further discussed in \cite{raissi_deep_2018}.
We select a set of $\Nf$ collocation points $\{ (x_j, t_j) \}_{j=1}^{\Nf}$, where we will link $\ut$ to $\Np$ through PDE residuals.
Unlike the samples of the state, these collocation points can be arbitrarily placed, and $\Nf$ can theoretically be taken as large as desired (subject to computational requirements).
To ensure that $\ut$ represents the noisy measurements and follows the partial differential equation, \cite{raissi_deep_2018} creates a compound loss function as
\begin{equation}
    \train =
    \frac{1}{N_u} \sum_{i=1}^{N_u} \left ( u_i - \ut (x_i, t_i) \right ) ^ 2 
    + \frac{1}{\Nf} \sum_{j=1}^{\Nf} \left ( \ut_t (x_j,t_j) - \left . \Np(\ut, \ut_x, \ut_{xx}) \right | _{(x_j,t_j)} \right ) ^ 2 .
    \label{eq:vanilla:loss}
\end{equation}
In the above, all derivatives of $\ut$ can be computed through automatic differentiation.
We can either train over $\theta$ first and then over $\phi$, or we can simultaneously minimize Eq. (\ref{eq:vanilla:loss}) over both $\theta$ and $\phi$.
The latter approach is highlighted further in Sec. \ref{sec:training:importance_of_simult}.
As previously mentioned, \cite{raissi_deep_2018} uses the former, staggered approach, first training $\ut$ to minimize Eq. (\ref{eq:mse}).
Then holding $\ut$ fixed, $\Np$ is trained to minimize the second term in Eq. (\ref{eq:vanilla:loss}).
Finally, $\ut$ is trained again on Eq. (\ref{eq:vanilla:loss}), now using the previously found $\Np$, with $\phi$ fixed.
This last step is identical to using a vanilla PINN \cite{raissi_physics-informed_2019} to solve the PDE given by $\Np$.
We describe this method of separately training $\theta$, then $\phi$, and then potentially $\theta$ as the ``non-simultaneous'' approach.

Despite the success of this strategy in recovering benchmark governing equations, there are a number of weaknesses, the first of which was originally highlighted in \cite{raissi_deep_2018}.
First, this method is very sensitive to noise in the data.
Dealing with noise is absolutely critical for working with experimental data.
Second, the unconstrained formulation in Eq. (\ref{eq:vanilla:loss}) enforces the identified equations in the form of a penalty; hence, there is no guarantee that the representation of the state $\ut$ follows the discovered PDE to high accuracy \cite{Basir2022_failure_modes, daw_rethinking_2022, wang_when_2022, Wang2022_respecting, wang_understanding_2021, krishnapriyan_characterizing_2021}.
Finally, the recovered neural network PDEs have not been solved with classical numerical methods (such as finite differences), which have extensive theoretical and historical support, unlike PINNs.
In the rest of this study, we aim to tackle these (somewhat related)  shortcomings.

\section{Methods}
\label{sec:methods}
Machine learning methods generally consider three regimes: training, validation, and testing.
Training is typically a gradient-based optimization problem over model parameters.
Validation is a similar nonlinear optimization problem, but over possibly discrete hyperparameters that may themselves control the gradient-based optimization \cite{raissi_open_2023}.
After all training and validation, testing then evaluates the model.
We similarly split our contributions into training and validation subsections.
The next subsection overviews two gradient-based training methods, then Sec. \ref{sec:validation} includes some further small, but important, implementation details.

\subsection{Training}
For our application, training should tune the parameters of the neural network describing the PDE, $\phi$, through also tuning the intermediate parameters $\theta$.
Similar to the method presented in Sec. \ref{sec:dhpm}, the training will not involve any PDE solves (with classical methods), and it does not deal with the selection of hyperparameters, which is discussed in Sec. \ref{sec:validation}.

Consider the two objectives represented in the compound loss function in Eq. (\ref{eq:vanilla:loss}).
The first objective is to match the potentially noisy data with our neural network representation of the state, $\ut$.
The second goal is to have this learned state satisfy our learned PDE, which is enforced at a finite set of collocation points.
If the data is noisy, we expect the data-matching objective to be nonzero, otherwise, we are overfitting to the noise.
Alternatively, satisfying the PDE is not directly influenced by the noise, and we can require this goal to be satisfied more strictly.
Thus, we pose the following constrained optimization problem as a reformulation of our two primary goals:

\begin{equation}
    \begin{aligned}
        & \min_{\theta, \phi} \quad \frac{1}{N_u} \sum_{i=1}^{N_u} \left (  u_i - u^\theta (x_i,t_i) \right )^2 \\
        & \text{subject to} \quad \underbrace{\ut_t(x_j,t_j) - \left . \Np(\ut, \ut_x, \ut_{xx}) \right | _{(x_j,t_j)}}_{\fj}  = 0, \quad j = 1,2,\hdots,\Nf.
    \end{aligned}
    \label{eq:general:base_constrained_form}
\end{equation} 
As shown in Eq. (\ref{eq:general:base_constrained_form}), we denote the PDE residual terms as $\fj$, noting these residual terms depend on the parameters of both the neural network approximation of the state, $\ut$, and on the neural network of the unknown PDE, $\Np$.

\subsubsection{Penalty-Method Formulation}
\label{sec:sa}
Several notable works in the PINNs literature include a similar constrained formulation to Eq. (\ref{eq:general:base_constrained_form}), but where $\Np$ is replaced with at least partially known physics \cite{son_-pinns_2023, Basir2022_PECANNs, lu_physics-informed_2021, hansen_learning_2023, xu_physics_2022}.
Specifically, the popular Self-Adaptive PINNs (SA PINNs) \cite{Mcclenny2022} consider the following, equivalent form, but for the specific case where the governing PDE is fully known ($\fj = r_{j}^{\theta}$),
\begin{equation}
      \min_{\theta, \phi} \, \max_{\lambda} \quad \frac{1}{N_u} \sum_{i=1}^{N_u} \left (  u_i - u^\theta (x_i,t_i) \right )^2 + \frac{1}{\Nf} \sum_{j=1}^{\Nf} \left ( \lambda_j \fj \right ) ^2   .
    \label{eq:sa:min_max_form}
\end{equation}
The SA PINNs algorithm is inspired by penalty methods and uses some optimizer (such as Adam \cite{kingma_adam_2014}) to iteratively update $\theta$ to minimize the objective, while simultaneously updating $\lambda$ to maximize the same objective.
Some training variations then fix the values of $\lambda$ and continue optimizing over $\theta$ to reach a final solution, potentially with a different optimizer (such as L--BFGS).

To generalize SA PINNs to the neural-network-based equation-discovery paradigm, we group the parameters describing the PDE, $\phi$, with the parameters for the representation of the state, $\theta$, as implied by Eq. (\ref{eq:sa:min_max_form}).
Note, however, that just like the generalization of vanilla PINNs to deep hidden physics models (\cite{raissi_physics-informed_2019, raissi_deep_2018}), we can still train these parameters separately if needed.
Before any training begins, the values of $\lambda$ are initialized according to some distribution.
Anecdotally, the shape of the distribution is of little importance, but the scale of the distribution seems to have a substantial influence on the quality of solutions found.
Thus, we also propose introducing a new hyperparameter $\lamo$ to initialize the residual weights in Eq. (\ref{eq:sa:min_max_form}).
Specifically, we initialize the weights independently and identically distributed (i.i.d.) according to a uniform distribution:
\begin{equation*}
    \lambda_j \stackrel{\text{iid}}{\sim} \mathcal{U} (0, \lamo), \quad j = 1,2,\hdots, \Nf .
\end{equation*}
As shown by the objective function in Eq. (\ref{eq:sa:min_max_form}), this hyperparameter is similar to the weighting of different terms in vanilla PINNs, but still allows for the weights to be updated during training, as it controls only the starting collocation weights.
An alternative to this parameter would be using a different learning rate for $\lambda$ as compared to $\theta$ and $\phi$ \cite{Mcclenny2022}.
We refer to this method as the ``penalty-method approach'' to differentiate its use from the original SA PINNs, although it is fundamentally a generalization to the neural-network-based equation-discovery paradigm.

\subsubsection{Barrier-Method Formulation}
\label{sec:constrained_method}
In this section, we refine the constrained optimization problem from Eq. (\ref{eq:general:base_constrained_form}), and introduce a more sophisticated solution of the constrained optimization than the penalty method that was presented in Sec. \ref{sec:sa}.
Given the unknown and complicated structure of neural networks, the constraints in Eq. (\ref{eq:general:base_constrained_form}) might not hold exactly for \textit{any} values of $\theta$ or $\phi$.
Thus, we propose an intermediate problem where these constraints are loosened as $| \fj | \leq \varepsilon$, for some small hyperparameter $\varepsilon$, yielding a new constrained optimization problem:
\begin{equation}
    \begin{aligned}
        & \min_{\theta, \phi} \quad \frac{1}{N_u} \sum_{i=1}^{N_u} \left (  u_i - u^\theta (x_i,t_i) \right )^2, \\
        & \text{subject to} \quad \fj -\varepsilon \leq 0, \quad j = 1,2,\hdots,\Nf, \\
        & \text{   and   } \quad -\fj -\varepsilon \leq 0, \quad j = 1,2,\hdots,\Nf.
    \end{aligned}
    \label{eq:constrained:loose_constrained_form}
\end{equation}

This constrained optimization problem is solved through a general, widely used, trust-region barrier method, which we briefly detail next.
Further details are presented in \ref{sec:sqp} and \cite{byrd_interior_1999}.
To convert the inequality constraints into equality constraints, this method adds slack variables $s_j, j = 1,2,\hdots, 2 \Nf$, as given below in Eq. (\ref{eq:constrained:constrained_form}). 
\begin{equation}
    \begin{aligned}
        & \min_{\theta, \phi, s} \quad \frac{1}{N_u} \sum_{i=1}^{N_u} \left (  u_i - u^\theta (x_i,t_i) \right )^2 - \mu \sum_{j'=1}^{2\Nf} \log s_{j'} \\
        & \text{subject to} \quad \fj -\varepsilon + s_j  = 0, \quad j = 1,2,\hdots,\Nf, \\
        & \text{   and   } \quad - \fj -\varepsilon + s_{j+\Nf} = 0, \quad j = 1,2,\hdots,\Nf.
    \end{aligned}
    \label{eq:constrained:constrained_form}
\end{equation}
Note that as the barrier parameter $\mu$ converges to $0$, the sequence of solutions to Eq. (\ref{eq:constrained:constrained_form}) should converge to a stationary point of Eq. (\ref{eq:constrained:loose_constrained_form}) \cite{byrd_interior_1999}.
A sequence of these equality-constrained optimization problems are solved for decreasing barrier parameter $\mu$ through a trust-region Sequential Quadratic Program (SQP) \cite{byrd_interior_1999, lalee_implementation_1998}.
It is worth highlighting that this optimization method usually requires 2nd-derivative information.
However, rather than computing expensive Hessians, we use a BFGS update to iteratively build approximations, based on \cite{nocedal_numerical_2006}, thus only using first derivative information.
For both the optimization procedure and these BFGS approximates, we use the implementations provided by SciPy \cite{virtanen_scipy_2020} in \texttt{scipy.optimize.minimize} with \texttt{method='trust-constr'}.
Despite the penalty-method formulation from Sec. \ref{sec:sa} also being related to a constrained formulation, we only refer to this trust-region barrier method as the ``constrained approach''.
We choose this general name, as this approach includes several sophisticated features beyond just the barrier method itself -- see \ref{sec:sqp} or \cite{byrd_interior_1999, lalee_implementation_1998, nocedal_numerical_2006}.

\subsubsection{Importance of Simultaneous $\Np$ and $\ut$ Training}
\label{sec:training:importance_of_simult}
We clarify one small, but important, training detail which differentiates this work from that in \cite{raissi_deep_2018}.
Here, we focus on the original formulation presented in Eq. (\ref{eq:vanilla:loss}), but this idea also applies to the penalty-method variant in Eq. (\ref{eq:sa:min_max_form}) and the constrained approach in Eq. (\ref{eq:constrained:loose_constrained_form}).

The ``non-simultaneous'' method first fits $\ut$ to the data through Eq. (\ref{eq:mse}).
Then, $\Np$ is trained on the full objective function in Eq. (\ref{eq:vanilla:loss}), (with the minimization only over only $\phi$).
Finally, to find a solution, $\ut$ is trained again, now incorporating the trained $\Np$, on Eq. (\ref{eq:vanilla:loss}) (or as discussed in the Sec. \ref{sec:solving_nn_pdes}, a standard PDE solver can be used).
This iterative training of $\ut$ then $\Np$ likely takes fewer, shorter training epochs, but it creates an increased sensitivity of $\Np$ to the noise in the data.
Fitting $\ut$ to data and computing derivatives through automatic differentiation is still somewhat sensitive to the underlying noise \cite{lagergren_learning_2020}.
Certainly, this global approach is preferred over other local \textit{a priori} methods \cite{cortiella_priori_2022, rudy_deep_2019}, but the performance is still poor.
Instead, simultaneously training both $\ut$ and $\Np$ on Eq. (\ref{eq:vanilla:loss}) (Eq. (\ref{eq:sa:min_max_form}) or Eq. (\ref{eq:constrained:loose_constrained_form})) regularizes this operation, promoting robustness of $\ut$ and its derivatives to noise.

Like many ill-conditioned operations, numerical differentiation can be improved through regularization \cite{Chartrand2011}.
We can regularize $\ut$ by adjusting the capacity of the neural network, applying early stopping, using data augmentation, or adding a regularization term to the loss function \cite{goodfellow_deep_2016, bishop_training_1995}.
The simultaneous training of $\ut$ and $\Np$ is regularization through a penalty in the loss function.
First, consider standard Tikhonov regularization, which adds a penalty $\Vert \theta \Vert _{2} ^2$ to the mean-squared error loss function (which is equivalent to weight decay with gradient descent \cite{krogh_simple_1991}).
The continuous generalization to functions instead adds $\Vert \ut \Vert_{L^2} ^2$ as a penalty, or even a Sobolev penalty involving $\ut_t$ and/or $\ut_x$ \cite{Chartrand2011}, where the integral must be approximated numerically.
For example, considering $H^1 (\Omega \times [0,T])$, the penalty term is
\begin{equation*}
    \int_0^T \int_\Omega \left (\ut(x,t) \right )^2 + \left ( \ut_x(x,t)  \right ) ^2 +\left ( \ut_t(x,t)  \right )^2 \ \mathrm{d} x \  \mathrm{d} t
    \approx \sum_{j=1}^{\Nf} \left ( \left (\ut(x_j,t_j) \right )^2 + \left ( \ut_x(x_j,t_j)  \right ) ^2 +\left ( \ut_t(x_j,t_j)  \right )^2 \right ) w_j ,
\end{equation*}
where $\{ (x_j, t_j) \}_{j=1}^{\Nf}$ are abscissas and $\{ w_j \}_{j=1}^{\Nf}$ are their associated weights.
A Monte Carlo approximation of this integral uniformly samples $(x_j, t_j)$ with weights $w_j = 1 / \left ( \Nf \cdot |\Omega| \cdot T \right )$ for $j=1,2,...,\Nf$.
This would promote a smoother $\ut$ used for the approximation of derivatives.
We can view the loss function in Eq. (\ref{eq:vanilla:loss}) as a further generalization to a non-Sobolev penalty.
Rather than generally promoting a smoother $\ut$, this penalty promotes the function $\ut$ satisfying the PDE $\ut_t = \Np (\ut, \ut_x, \ut_{xx}, ...)$, in the $L^2$ sense
\begin{equation*}
    T\cdot |\Omega| \cdot \int_0^T \int_\Omega \left ( \ut_t (x,t) -  \Np (\ut, \ut_x, \ut_{xx}, ...) \right )^2 \ \mathrm{d} x \  \mathrm{d} t
    \approx \frac{1}{\Nf} \sum_{j=1}^{\Nf} \left ( \ut_t (x,t) -  \Np (\ut, \ut_x, \ut_{xx}, ...) \right )^2 .
\end{equation*}
This does not necessarily promote a smoother $\ut$, but depending on $\Np$ it may (take $\Np = 0$ or $\Np = \ut_x$ for instance).
Without loss of generality, higher-order terms could be added, as above, and as in the PINNs literature \cite{gpinns}.

Thus, simultaneous training on Eq. (\ref{eq:vanilla:loss}) adds regularization to the training of $\ut$, causing less sensitivity to noise in the data.
Of course, $\Np$ also updates during this process, changing the specifics on how $\ut$ is regularized at each step.
Similar to standard regularization techniques, we often scale this term by some factor.
In the penalty method, $\lambda_0$ serves this role, now with the $\lambda_{j}^2$ terms both scaling the entire regularization term and acting as integration weights $w_j$.
Similarly, in the constrained method, decreasing $\varepsilon$ increases the strength of this regularization.
In Sec. \ref{sec:burg:nl} we make a mention to the possible improvement provided by this simultaneous training.

\subsection{Validation and Other Implementation Details}
\label{sec:validation}

In the penalty-method approach and in the constrained approach, we introduced new hyperparameters, $\lamo$ and $\varepsilon$, which are related to the PDE residuals at the collocation points, $\fj$.
A larger initial penalty weight $\lamo$ prioritizes the $\fj$ terms to be decreased first (Eq. (\ref{eq:sa:min_max_form})).
Conversely, a smaller constraint looseness parameter $\varepsilon$, requires that the $\fj$ terms must be closer to zero.
Thus, as explained in Sec. \ref{sec:solving_nn_pdes}, $1 / \lamo$ plays a similar role to $\varepsilon$.
Next, we will discuss how these hyperparameters may be tuned using a validation loss.

For time series data, there are multiple ways to split a dataset into training and validation sets \cite{cerqueira_evaluating_2020}.
We will use the simple approach of having only one training / validation split, where all the validation data occurs later in time than the training data.
We will use all the available data, meaning that there is no window between the training and validation datasets.

Further, for all of the methods we have presented, we have multiple notions of the state for any given data point $((x_i, t_i), u_i)$.
Most obviously, we have the neural network approximation of the state $\ut (x_i, t_i)$.
More subtly, the learned PDE $\Np$ also provides information about the state, when paired with the relevant initial condition and boundary condition information.
That is, (numerically) solving this PDE gives another approximation of the state, which we use for validation, without using it in training.
The approximation $\ut (x_i,t_i)$ can itself be considered as a solution of the learned PDE $\Np$, when using a PINNs-like numerical method, which also incorporates the available data, and this is the notion used during training. We are mostly concerned with using a traditional numerical PDE solver to compute solutions of $\Np$, unlike in \cite{raissi_deep_2018}, where the discovered PDEs are solved through PINNs \cite{raissi_physics-informed_2019}.
In this work, we prioritize using a traditional numerical PDE solver to compute solutions of $\Np$, in contrast to the approach in \cite{raissi_deep_2018}, which uses PINNs \cite{raissi_physics-informed_2019} to solve the discovered PDEs.
We focus on classical solvers because, despite their potential, PINNs still face significant challenges in solving forward PDEs, even for relatively simple cases \cite{chuang_experience_2022}. 
The PDE residual term in PINNs creates a complex loss surface that is difficult for optimizers to navigate \cite{krishnapriyan_characterizing_2021}.
Additionally, the placement and weighting of these collocation points remains an active area of research
\cite{amin_nabian_efficient_2021, daw_rethinking_2022, basir_investigating_2023, Wang2022_respecting, wang_understanding_2021, wang_when_2022}, and the incorporation of initial and boundary conditions is not straightforward \cite{lu_physics-informed_2021, bischof_multi-objective_2022}.
Unlike classical methods, PINNs lack many of the established theoretical guarantees \cite{cuomo_scientific_2022}.
While we acknowledge that PINNs are likely to improve, these current limitations lead us to favor classical numerical schemes. Thus, we will rely on a classical numerical solution to our neural network PDE for evaluating both the validation and testing errors.

\subsubsection{Solving neural network PDEs}
\label{sec:solving_nn_pdes}
We use the method of lines to solve the learned neural network PDE $\Np$.
Here we will assume that $\mathcal{N}^\phi = \mathcal{N}^\phi(x,t,u,u_x,u_{xx})$, without loss of generality.
As outlined in Sec. \ref{sec:background:mol}, the method of lines discretizes the PDE into a system of ODEs on $n_x - 1$ interior grid points $\{ x_k \}_{k=1}^{n_x - 1}$.
The operator $\Np$, paired with discretizations of the spatial derivatives and the boundary conditions, gives the evolution operator of this system of ODEs (see Eq. (\ref{eq:method_of_lines})).
With $\Np$ as a neural network, the integration of the ODEs is similar to the work of Neural ODEs \cite{chen_neural_2019}, but where we also add the spatial discretization and boundary conditions as an intermediate step.
This system of ODEs is solved through a time integration scheme, which we take to be equi-spaced with $n_t$ total times, excluding the initial condition.
Generally, we determine $n_t$ from the number of spatial points, and thus just change $n_x$ to both refine  the spatial mesh and to decrease the time step.
Including boundary conditions and the initial condition, with the temporal discretization $\{ t_l \}_{l=0}^{n_t}$, this gives a solution of the form
\begin{equation}
    U_{kl}^{\phi} = u^{\phi, n_x} (x_k, t_l) \approx u(x_k, t_l).
    \label{eq:solving_nn_pdes}
\end{equation}
Note here that the solution over spatial and temporal discretization, $U_{kl}^{\phi}$, depends on the PDE we are solving and thus also on its parameters $\phi$, but it does \textit{not} depend on $\theta$, the parameters for the denoised state representation $\ut$.
Here, we also include an explicit reference to the number of spatial mesh points $n_x$, to clarify our later use of multiple spatial meshes.

However, the validation data does not exist on this same grid as this PDE solution, instead requiring us to interpolate the gridded PDE solution to give predictions at the location of the validation data.
In practice, with a reasonably fine mesh, interpolation errors are insignificant, and thus, we simply use a bilinear interpolant for all interpolations.
For a point $(x,t)$, we denote this interpolated solution as $u^{\phi, n_x} (x,t)$. The validation loss, \val, is then computed through the mean squared error against the validation dataset $\{ (x_{i'}, t_{i'}), u_{i'} \}_{{i'}=1}^{N_{u}'}$, consisting of $N_{u}'$ points, occurring after the latest training point ($t_{i'} \geq t_i, \forall i, \forall i')$. Note, because this data is unstructured, we index it with a single index, as opposed to the meshed solution in Eq. (\ref{eq:solving_nn_pdes}) which is indexed separately in space and time. With this in mind, the validation loss is
\begin{equation}
    \val = \frac{1}{N_{u}'} \sum_{{i'}=1}^{N_{u}'} \left ( u_{i'} - u^{\phi, n_x} (x_{i'}, t_{i'}) \right )^2 .
    \label{eq:validation}
\end{equation}

As previously mentioned, the solution $u^{\phi, n_x}$ comes from applying the method of lines and a time integrator to solve the neural network PDE, $\ut = \Np(u, u_x, u_{xx}, ... )$.
Because neural networks are generally highly nonlinear, noninterpretable, and poorly understood, our numerical method may have some additional difficulties.
It is worth also noting that the training process itself does \textit{not} rely on the classical numerical solutions to the PDE that we are now considering.
In particular, one should consider the details of the numerical solver, including the following: the time integration method, the number of time steps $n_t$, the numerical approximation of the spatial derivatives, and the number of grid points $n_x$, but
such an analysis is beyond the scope of this work.
However, from our experience and on our example problems, the time integration method and the number of time steps seem to be less significant than the choice of the spatial derivatives and the number of grid points.
For instance, increasing $n_t$ by a factor of 10 generally results in no change to the solution, but in some cases, doubling $n_x$ changes the numerical solution.
To deal with this spatial discretization issue, without making it the focus of this work, we will use multiple mesh resolutions when computing the validation loss.
This promotes more mesh-independent behavior of the PDE operator we select.
We compute $\val$ with three meshes, where the number of mesh points are given by $n_x = n_{x}^{1}$, $n_{x}^{2}$ and $n_{x}^{3}$.
We use the simple approach of computing multiple values of $\val$, where we change the number of grid points, $n_x$, used to discretize the PDE in space.
For each change to $n_x$, we adjust the time step so that the ratio between the time step and the spatial mesh size remains the same.
We keep the same time integration method and the same spatial differentiation method.
For this paper, we use the classic 4th-order Runge-Kutta and centered finite differences.
We report the final validation loss $L_\text{val} (\phi)$ as the maximum mean-squared error (as defined in Eq. (\ref{eq:validation})) over these three meshes.
That is,
\begin{equation}
    L_\text{val} = \max \ \left \{ \  L_{\text{val}}^{n_{x}^1}, L_{\text{val}}^{n_{x}^2}, L_{\text{val}}^{n_{x}^3 } \  \right \} .
    \label{eq:validation:final}
\end{equation}
This modified validation loss definition aims to provide some robustness to the mesh used for the numerical solution of the PDE $\Np$.
Anecdotally, increasing the number of collocation points also improves the robustness of the numerical solutions.
This method is only a temporary fix, and future work may focus on a more refined approach, such as data-driven discretizations which provide stability guarantees \cite{prakash_data-driven_2024}.

\subsubsection{PDE Importance Parameter ($\varepsilon$ or $\lamo$)}
In the penalty and constrained approaches, we have made mention to hyperparameters $\lamo$ and $\varepsilon$, which control the trade-off between matching the data and driving the PDE residuals to zero.
In Alg. \ref{alg:train_validation}, we propose to use the validation loss from Eq. (\ref{eq:validation:final}) to select these hyperparameters.
We also use this validation process to select between different random initializations of the $\Np$ and $\ut$ networks.

\begin{algorithm}
    \begin{algorithmic}[1]
        \STATE Validation method for selecting $\varepsilon$ or $\lamo$ and the neural network seed
        \FOR{neural network seed $s= 1,2,3$}
            \FOR{$k = 1,2, \hdots, 10$}
                \STATE \textbf{Initialize} $(\ut)^{k,s}$ and $(\Np)^{k,s}$ with parameters $\theta^{k,s}$ and $\phi^{k,s}$ using neural network seed $s$
                \STATE \textbf{Train} $\theta^{k,s}$ and $\phi^{k,s}$ according to Eq. (\ref{eq:constrained:loose_constrained_form}) or Eq. (\ref{eq:sa:min_max_form}), using hyperparameter $\varepsilon^{k,s}$ or $(\lamo)^{k,s}$, respectively
                \STATE \textbf{Compute} PDE solution $(u^\phi)^{k,s}$, using Method of Lines with $\Nc = (\Np)^{k,s}$ (See Eq. (\ref{eq:method_of_lines}) and Eq. (\ref{eq:solving_nn_pdes})) 
                \STATE \textbf{Compute} $L_{\text{val}}^k = L_{\text{val}} (\phi^{k,s})$, which uses $(u^\phi)^{k,s}$ (See Eq. (\ref{eq:validation}))
            \ENDFOR
            \STATE \textbf{Select} $K = \argminop_k \ L_{\text{val}}^{k,s}$
        \ENDFOR
        \STATE \textbf{Select} $S = \argminop_s \ L_{\text{val}}^{K,s}$
        \RETURN $(\Np)^{K,S}$
    \end{algorithmic}
    \caption{Hyperparameter Selection through Method of Lines Validation}
    \label{alg:train_validation}
\end{algorithm}

\subsection{Testing}
The training and validation in Alg. \ref{alg:train_validation} use data from the system of interest, $\{ (x_i, t_i), u_i \}_{i=1}^{N_u}$ and $\{ (x_{i'}, t_{i'}), u_{i'} \}_{{i'}=1}^{N_{u}'}$, to give a final learned PDE $\Np$.
Specifically, $u_i$ and $u_{i'}$ are samples of the system, for a single initial condition.
To test the effectiveness of the learned PDE $\Np$, we will test how well it extrapolates to a new, unseen initial condition.
We refer to this as ``testing'', although in our case we are applying our model $\Np$ outside the training/validation distribution.
We will refer to the initial condition used to generate the training/validation data as \utrain, and refer to the unseen initial condition as \utest.
A different measure of performance could test the PDE on a different geometry or with different boundary conditions than those used in training/validation, but we limit our tests to changing the initial condition.

\section{Results}
\label{sec:results}

We now proceed to some numerical examples illustrating the methods.
To generate a true baseline solution, we numerically solve each known PDE with a Fourier spectral method, as in \cite{raissi_deep_2018}, denoting this solution $U_{kl}^{\text{true}}$ on the spatial-temporal grid $\{ (x_k, t_l) \}_{k=0, l=0}^{k=n_x+1, l=n_t}$ (for both \utrain and \utest).
Then to generate the training data, we add independent, identically distributed noise to the training and validation data, which both come from the \utrain initial condition.
Again, as done in \cite{raissi_deep_2018}, this noise is normally distributed, with standard deviation given by a $\text{noise level}$ parameter times a scaling factor based on the true solution ($\mathrm{std}_{k,l} \, U_{kl}^{\text{true}}$).
Specifically, we draw noise instance $\eta_{kl}$ independently, identically normally distributed with mean $0$ and standard deviation
\begin{equation*}
\text{noise level} \ \cdot \ \mathrm{std}_{k,l} \, U_{kl}^{\text{true}} .    
\end{equation*}

To evaluate the performance of the methods, we will use a relative $\ell^2$ error.
We wish to evaluate the performance of the (interpolated) numerical solution of $\Np$, which is denoted $\up$, as explained in Sec. \ref{sec:solving_nn_pdes}.
The error metric is explicitly defined as
\begin{equation}
    \rell (\phi) = \sqrt { \frac{ \sum_{k=0}^{n_x+1} \sum_{l=0}^{n_t}  \left (U_{kl}^{\text{true}} - \up (x_k, t_l) \right )^2 }
    {\sum_{k=0}^{n_x+1} \sum_{l=0}^{n_t}  \left (U_{kl}^{\text{true}} \right )^2} } .
    \label{eq:results:rell}
\end{equation}
This metric is computed for both \utrain and \utest, by appropriately changing $U_{kl}^{\text{true}}$ and the initial condition used in the solution of $\Np$.
Note that this metric is quite different from both the training term $L_\text{MSE}$ (Eq. (\ref{eq:mse})) and the validation loss $L_\text{val}$ (Eq. (\ref{eq:validation})), as it is computed over a grid and uses noiseless data.
Further, it combines both the training and validation domains into one metric. That is,
\begin{align*}
    \{ (x_i, t_i) \}_{i=1}^{N_u} &\subset \{ (x_k, t_l) \}_{k=0, l=0}^{k=n_x+1, l=n_t}\\
    \{ (x_{i'}, t_{i'}) \}_{{i'}=1}^{N_{u}'} &\subset \{ (x_k, t_l) \}_{k=0, l=0}^{k=n_x+1, l=n_t} .
\end{align*}

For large values of $\rell$, Eq. (\ref{eq:results:rell}) may not be a helpful measure of performance.
Thus, we compute another metric, the time-to-failure.
This metric describes how long the solution of the discovered PDE takes to diverge from the solution of the true PDE.
With this intuition, a greater time-to-failure indicates better agreement between the discovered and true PDEs.
We define the time-to-failure as the earliest time where the spatial, relative error of the numerical solution of $\Np$ exceeds some threshold, $\delta$.
If the relative error of the numerical solution of $\Np$ never exceeds the threshold, then we take the time-to-failure as the total simulation time, $T$.
As above, these errors are taken in the $\ell^2$ norm.
In other words, if the time-to-failure is $t_l < T$, then
\begin{equation}
    \frac{\norm{U_{\bullet l}^{\text{true}} - u^\phi ( \bullet, t_l ) }_2 }
    {\norm{U_{\bullet l}^{\text{true}}}_2} > \delta ,
\label{eq:results:ttf}
\end{equation}
and there is no earlier time where the above would hold. If the time-to-failure is $T$, there is no such time $t_l$ where the above holds.

\subsection{Burgers' Equation}
\label{sec:burgers}
We consider the following system governed by Burgers' Equation:
\begin{equation}
    \begin{aligned}
        &u_t = -u u_x - 0.1 u_{xx}, & -8 &< x < 8, \ 0 < t < T, \\
        &u(x,0) = u_0(x), & -8 &\leq x \leq 8, \\
        &u(-8,t) = 0, & 0 &\leq t \leq T, \\
        &u(8,t) = 0, & 0 &\leq t \leq T.
    \end{aligned}
    \label{eq:burgers}
\end{equation}
Again, we use different versions of $u_0 (x)$ for training versus testing, which are given as
\begin{equation}
    \begin{aligned}
        &\utrain (x) = - \sin \left ( \frac{\pi x}{8} \right ),\\
        &\utest (x) = \exp{ \left ( - (x + 2)^2 \right ) }.
    \end{aligned}
    \label{eq:burgers:ics}
\end{equation}
For training, we use $T = 30$, and for testing, we use $T=10$.
In the training/validation dataset, we use $N_u = 10000$ randomly sampled points, taking the first two-thirds for training and using later third for the selection of hyperparameters during validation.
The true solution is saved on a mesh of $n_x = 256$ points, with $n_t = 600$ time steps for training and $n_t = 200$ time steps for testing.
We use $10$ instances of the $\varepsilon$ or $\lambda$ hyperparameters, selecting the best model based on the validation loss computed via Eq. (\ref{eq:validation}).
We also repeat this $3$ times with different neural network initializations, again choosing the best model through the validation loss.
This process is outlined in Alg. \ref{alg:train_validation}.
Next, we will vary the noise level and the number of collocation points $\Nf$ to evaluate the different methods.
The PDE network $\Np (u, u_x, u_{xx})$ is parameterized with 2 hidden layers, each with 16 units, with sine activation functions, and the intermediate denoised state representation $\ut (x,t)$ is a neural network with 5 hidden layers, each 32 units wide, using sine activation functions, inspired by SIREN \cite{sitzmann_implicit_2020}.
Again, we highlight that we use a classical numerical method to solve the learned PDEs, as opposed to using PINNs \cite{raissi_deep_2018, raissi_physics-informed_2019}.
The majority of the implementation is in PyTorch \cite{PyTorch}, with the codes available on GitHub (upon acceptance).
While we could use the adjoint method for computing gradients \cite{chen_neural_2019}, we found that this decreased speed, and we therefore use standard backpropagation throughout.

\subsubsection{Noise Level Study}
\label{sec:burg:nl}
For a given dataset, Alg. \ref{alg:train_validation} gives one prediction for the governing PDE as $\Np$.
We evaluate the performance of the discovered PDE by comparing its solution on the training initial condition to the true solution.
Similarly, we further scrutinize $\Np$ with the same comparison, but for the unseen testing initial condition, $\utest$.
This gives two error metrics from Eq. (\ref{eq:results:rell}).

\begin{figure}[htb!]
    \centering
    \includegraphics[width=0.85\linewidth]{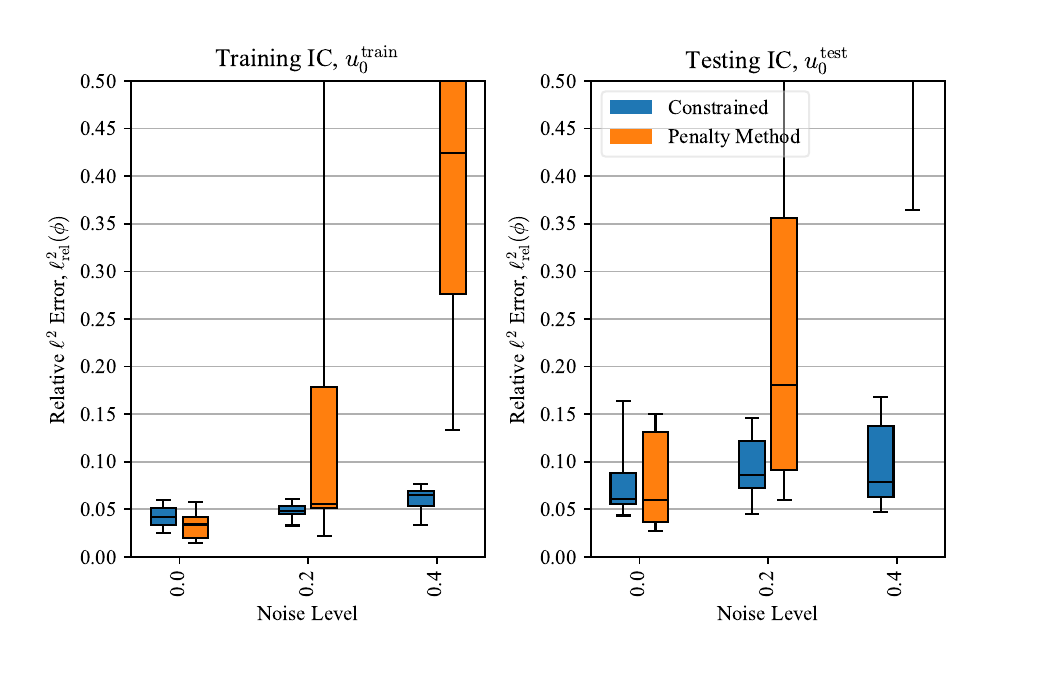}
    \caption{Using $\Nf = 1000$ collocation points, the proposed constrained formulation outperforms the penalty-method approach, for nonzero noise levels. On a very similar problem, \protect\cite{raissi_deep_2018} reported an error of $0.46$ for $\text{noise level} = 0.05$ for $\utrain$ (see Table 1). Recall that $\utest$ is entirely unseen during training. Each box-and-whisker shows ten datasets / initializations and runs through Alg. \ref{alg:train_validation} with validation. In the 0.0 noise level case, the location of the data, the location of the collocation points, and the neural network initializations change (and the initialization of $\lambda_j$ for the penalty-method approach).}
    \label{fig:burgers:noise_level:l2}
\end{figure}

We repeat training for 6 different scenarios, where the two training methods, i.e. the penalty-method and constrained approaches, were applied for three different noise levels ($0, 0.2, 0.4$).
For each of these scenarios, we repeat this for 10 initializations, changing the randomly sampled data locations, the collocation point locations, and the initialization of the collocation weights (for the penalty method).
This gives 60 repeats of Alg. \ref{alg:train_validation}, shown in Fig. \ref{fig:burgers:noise_level:l2}.
This ensemble approach gives a better overview of the performance of the training methods, as PINNs-like methods have been known to be inconsistent in some cases \cite{chuang_experience_2022, krishnapriyan_characterizing_2021}.
Considering that Alg. \ref{alg:train_validation} itself trains 30 neural networks, this is a substantial amount of computation -- hundreds or a few thousands of core-hours for the penalty-method approach or the constrained approach, respectively.

\begin{figure}[htb!]
    \centering
    \includegraphics[width=0.85\linewidth]{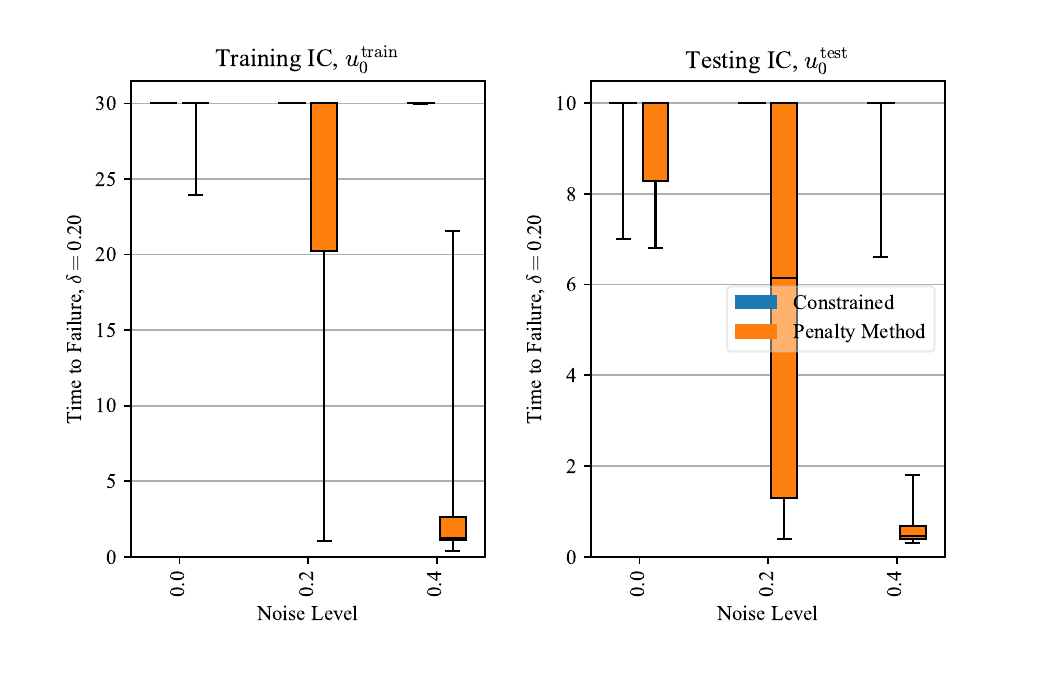}
    \caption{With the same setup and runs as Fig. \ref{fig:burgers:noise_level:l2}, we show the time-to-failure (Eq. (\ref{eq:results:ttf})). Higher is better for this performance measure. Note that for $\utrain$, $T=30$, but for $\utest$, $T = 10$.}
    \label{fig:burgers:noise_level:ttf}
\end{figure}

In Fig. \ref{fig:burgers:noise_level:l2} we hold the number of collocation points constant and evaluate the performance of the methods when the training and validation data are corrupted with increasing levels of noise.
For the proposed constrained formulation, we see almost constant performance across the noise levels, as opposed to the penalty method (Eq. (\ref{eq:sa:min_max_form})) where the performance degrades at the relatively-high noise levels. In this work, each box-and-whisker plot shows the minimum, first quartile (25th percentile), median, third quartile (75th percentile), and maximum values. From the variation in the noise-free case, we see that both methods are still considerably sensitive to the random initializations.
The $\rell$ varies by about 5\% on the training initial condition, and by more than 10\% on the extrapolation initial condition.
We suspect that the ill-posedness of the problem causes these fluctuations.

Fig. \ref{fig:burgers:noise_level:ttf} evaluates a different performance measure (Eq. (\ref{eq:results:ttf})) but tells the same story -- the constrained approach performs better than the penalty method for higher noise levels.
It generally takes longer for solutions from the constrained approach to exceed the tolerance.
For this problem, the constrained method usually discovers PDEs such that the solutions do not exceed the error tolerance, $\delta = 0.2$, during our simulation time $0 \leq t \leq T$, for both the training and testing initial conditions.

All the metrics reported in Figs. \ref{fig:burgers:noise_level:l2} and \ref{fig:burgers:noise_level:ttf} are computed by solving the learned $\Np$ on an equispaced mesh of $n_x = 128$ points, with $dt = 0.2 \cdot dx$, using 2nd-order centered finite differences over $x$ and an explicit 4th-order Runge-Kutta scheme in time.
As given by Eq. (\ref{eq:validation:final}) in Sec. \ref{sec:solving_nn_pdes}, the validation loss uses solutions from multiple meshes.
These meshes are $n_{x}^{1} = 112$, $n_{x}^{2} = 128$, and $n_{x}^{3} = 148$, selected arbitrarily around the testing mesh of $n_x = 128$.
The spatial derivatives are calculated through 2nd-order centered finite difference methods, each using just 3 points and thus only requiring the Dirichlet boundary condition information given in Eq. (\ref{eq:burgers}).
We use the same explicit 4th-order Runge-Kutta integrator, with $dt$ based on $dx$, through the same ratio as above.

\begin{figure}[htb!]
    \centering
    \includegraphics[width=0.85\linewidth]{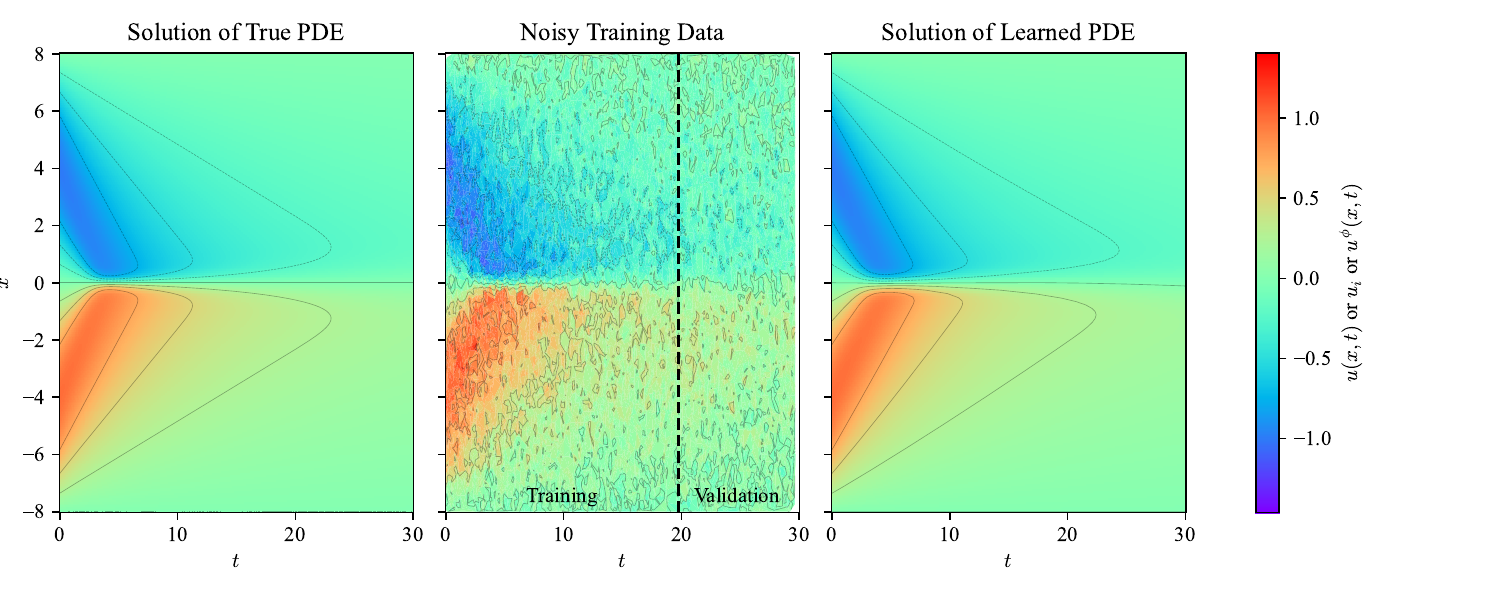}
    \caption{Despite noisy training data and relatively few collocation points, the constrained method accurately recovers the governing PDE. We show the solution of PDE discovered by the constrained method, for the highest noise level ($0.4$), with the lowest training error in Fig. \ref{fig:burgers:noise_level:l2}. This is the PDE solution for the $\utrain$ initial condition. The $\utest$ initial condition for the same discovered PDE is shown in Fig. \ref{fig:burgers:example:con:extrap}.}
    \label{fig:burgers:example:con:train}
\end{figure}

\begin{figure}[htb!]
    \centering
    \includegraphics[width=0.85\linewidth]{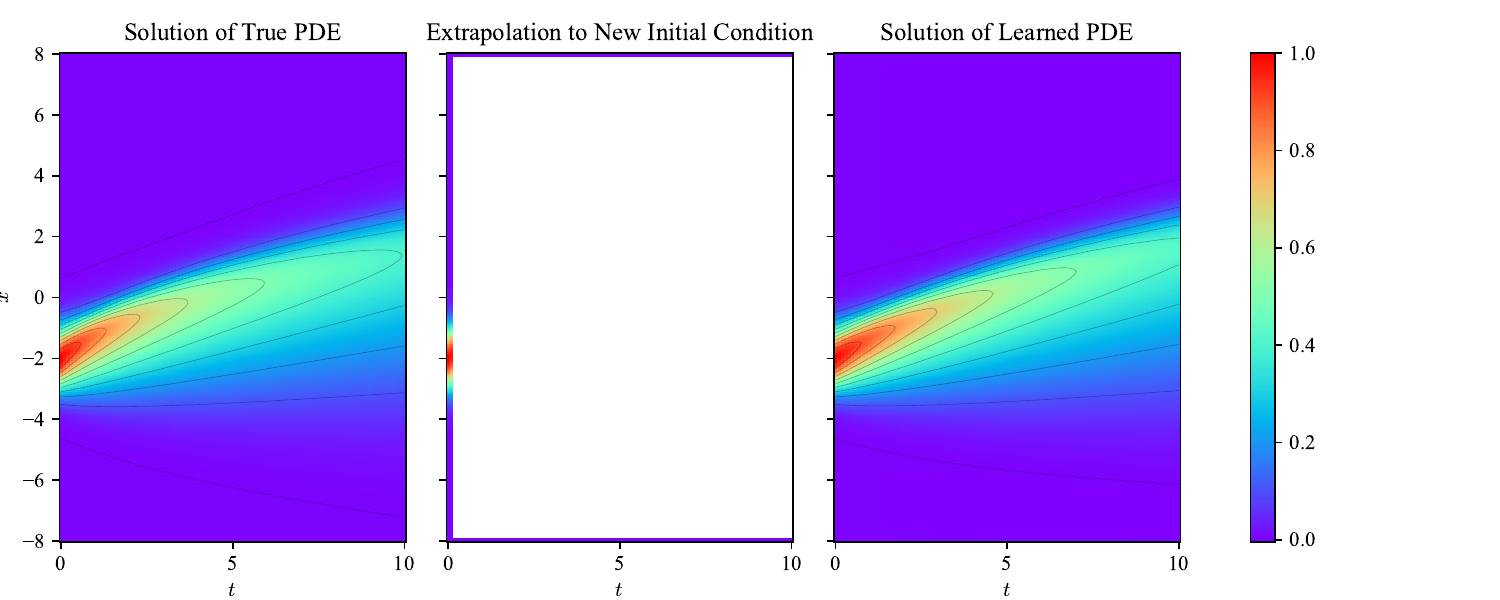}
    \caption{For the constrained method, the discovered PDE accurately predicts the evolution of an unseen initial condition. The PDE discovered in Fig. \ref{fig:burgers:example:con:train} is solved again with a classical finite difference method, but for $\utest$, while all of the (noisy) training data came originally from $\utrain$. In other words, the noisy data from $\utrain$ (middle plot of Fig. \ref{fig:burgers:example:con:train}) is used to train $\Np$, and the initial and boundary conditions for the new $\utest$ (middle plot above), are used for the numerical solution in the right plot above. This run corresponds to $\rell = 0.06$ for $\utest$ in Fig. \ref{fig:burgers:noise_level:l2}.}
    \label{fig:burgers:example:con:extrap}
\end{figure}

Figs. \ref{fig:burgers:example:con:train} and \ref{fig:burgers:example:con:extrap} show an example result from this noise level study, representing one iteration through Alg. \ref{alg:train_validation}.
Fig. \ref{fig:burgers:example:con:train} illustrates the magnitude of the $0.4$ noise level, and compares the classical numerical solution of the learned PDE $\Np$ to the solution of the true PDE.
More impressively, Fig. \ref{fig:burgers:example:con:extrap} shows the solution of this learned PDE $\Np$ on an entirely new initial condition.
Compared to the example presented in \cite{raissi_deep_2018}, this example uses 10 times fewer collocation points, a longer simulation time for the training initial condition, larger noise levels, and a classical numerical solver.
We highlight further comparison details in Sec. \ref{sec:burgers:nf} and Sec. \ref{sec:discussion:perf}.


\begin{figure}[htb!]
    \centering
    \includegraphics[width=0.85\linewidth]{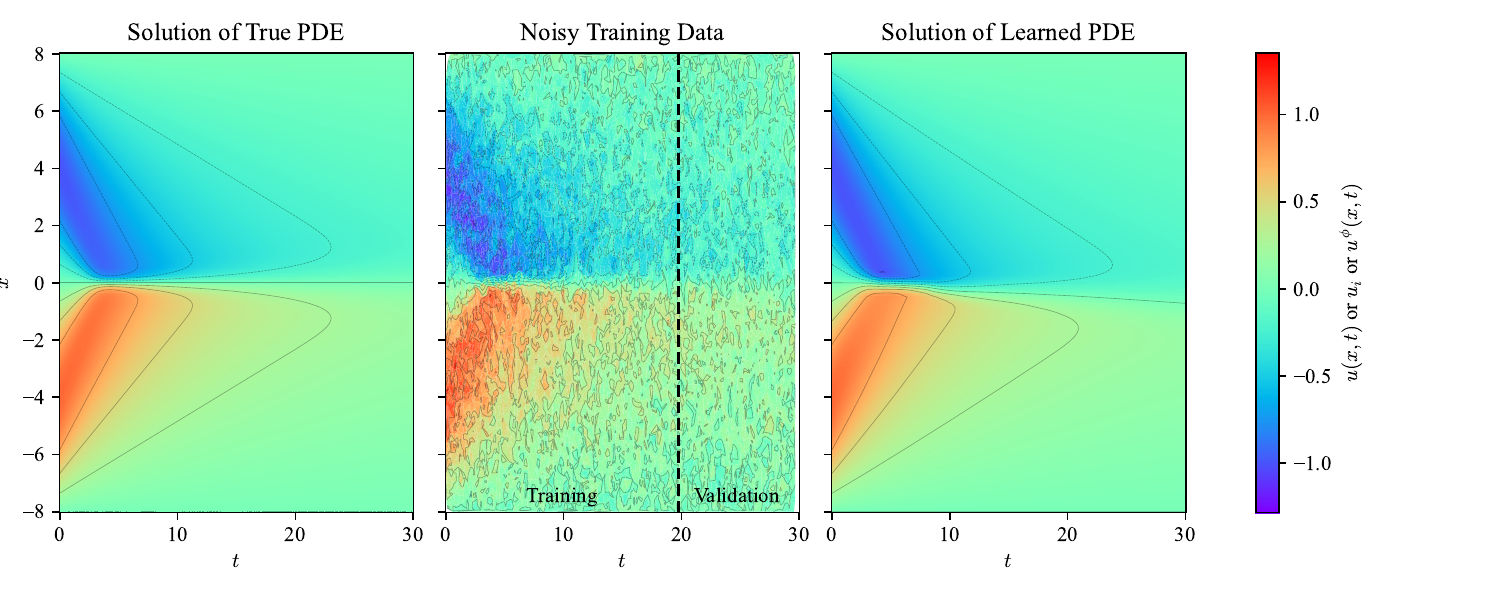}
    \caption{Similar to Fig. \ref{fig:burgers:example:con:train}, we show the solution of PDE discovered through the penalty-method approach, for the highest noise level ($0.4$), with the lowest training error in Fig. \ref{fig:burgers:noise_level:l2}. This is the PDE solution for the $\utrain$ initial condition. The $\utest$ initial condition for the same discovered PDE is shown in Fig. \ref{fig:burgers:example:sa:extrap}. Here, we see some asymmetry, with the $u(x,t) = 0$ contour slightly drifting from the $x=0$ axis.}
    \label{fig:burgers:example:sa:train}
\end{figure}
\begin{figure}[htb!]
    \centering
    \includegraphics[width=0.85\linewidth]{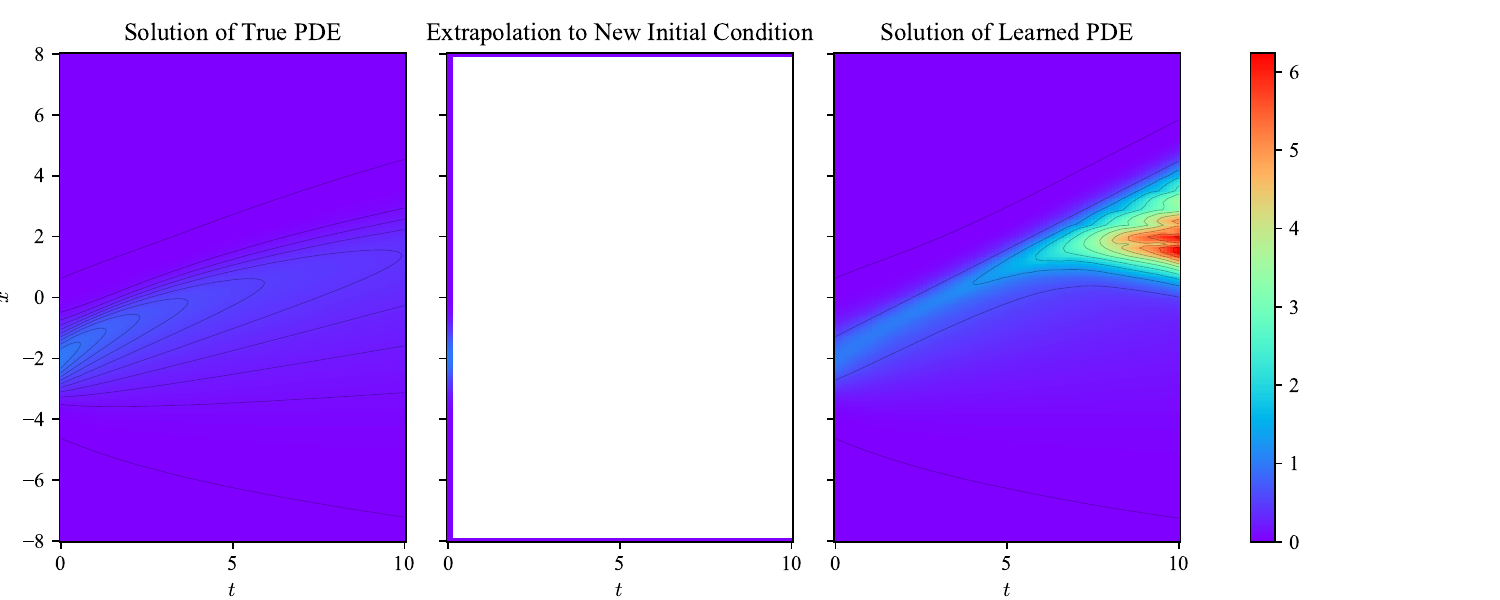}
    \caption{For the penalty-method approach, this discovered PDE fails to accurately predict the evolution of an unseen initial condition (as opposed to Fig. \ref{fig:burgers:example:con:extrap}). The PDE discovered in Fig. \ref{fig:burgers:example:sa:train}, is solved again with a classical finite difference method, but for $\utest$. Despite having a successful performance on $\utrain$, this discovered PDE does not generalize well to the unseen initial condition $\utest$. This may be due to the fundamental ill-posedness of the problem, rather than the specific method used.}
    \label{fig:burgers:example:sa:extrap}
\end{figure}

\subsubsection{Number of Collocation Points, $\Nf$}
\label{sec:burgers:nf}
\begin{figure}[htb!]
    \centering
    \includegraphics[width=0.85\linewidth]{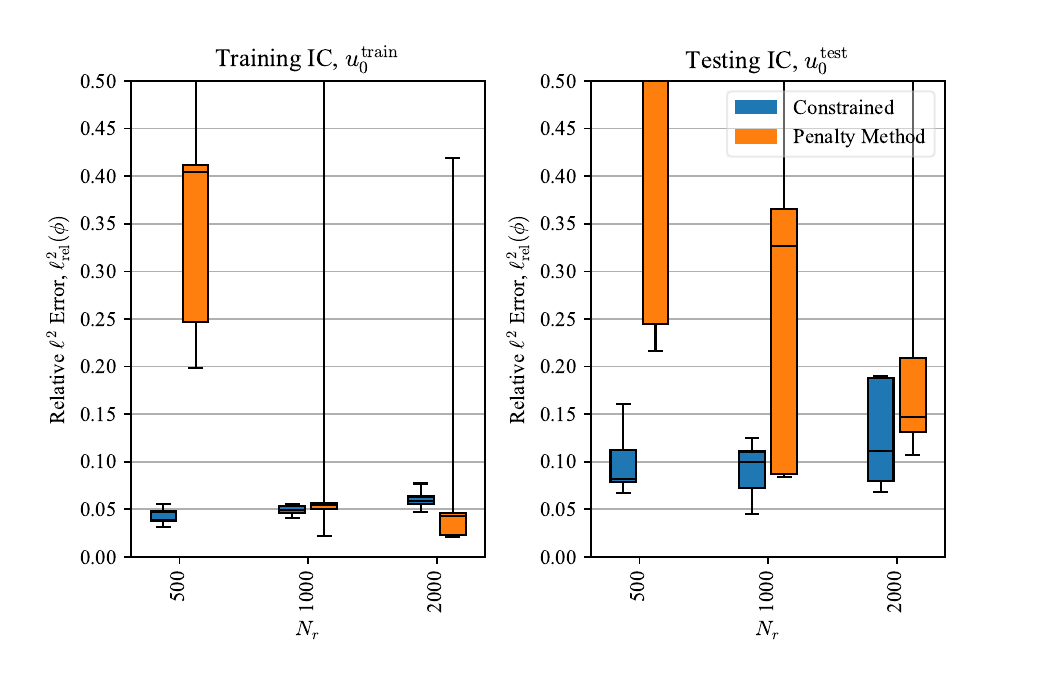}
    \caption{Particularly for fewer collocation points, the constrained method outperforms the penalty-method approach. These results are for the middle noise level of $0.2$, and each box-and-whisker plot is made up of 5 runs from Alg. \ref{alg:train_validation}, each applied to a different noise realization. Thus, each of the 5 runs is shown by a different feature of the box-and-whisker plot.
    }
    \label{fig:burgers:nf:l2}
\end{figure}

Here, we investigate how these methods depend on the number of collocation points, $\Nf$, again for the system given by Eq. (\ref{eq:burgers}) with the same two initial conditions from Eq. (\ref{eq:burgers:ics}). Other than $\Nf$, all other hyperparameters are kept the same.
As opposed to Fig. \ref{fig:burgers:noise_level:l2}, the results in Fig. \ref{fig:burgers:nf:l2} include five, rather than ten repetitions of Alg. \ref{alg:train_validation} (for each box-and-whisker).

\begin{figure}[htb!]
    \centering
    \includegraphics[width=0.85\linewidth]{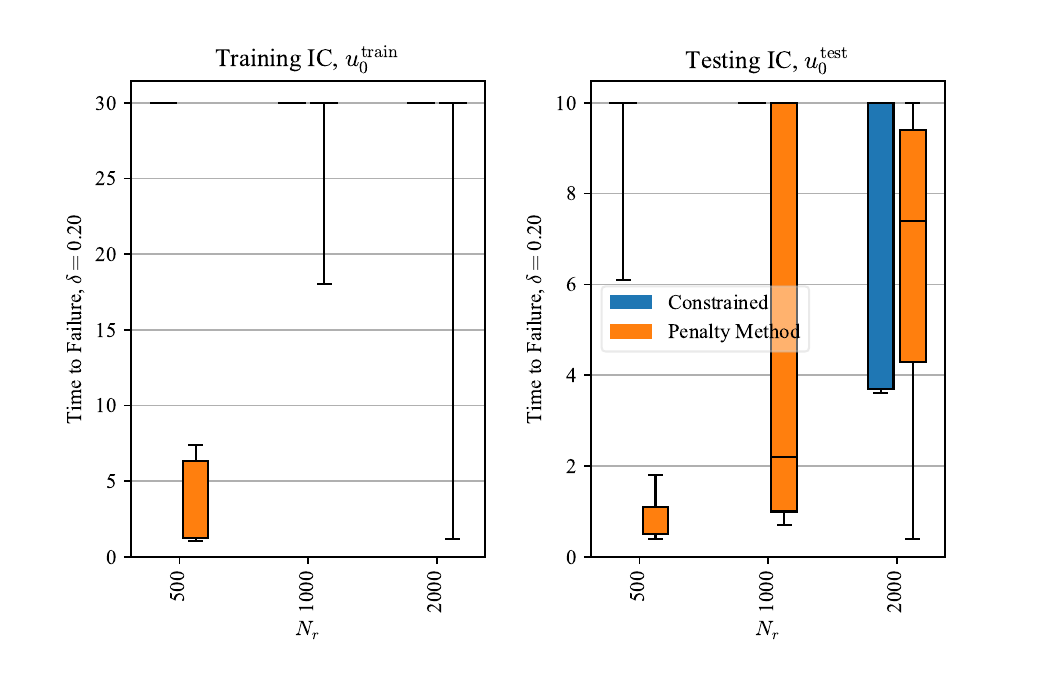}
    \caption{With the same setup and runs as Fig. \ref{fig:burgers:nf:l2}, we show the time-to-failure (Eq. (\ref{eq:results:ttf})). Higher is better for this performance measure. Note that for $\utrain$, $T=30$, but for $\utest$, $T = 10$.}
    \label{fig:burgers:nf:ttf}
\end{figure}

Fig. \ref{fig:burgers:nf:l2} shows the $\rell$ error measure for the two initial conditions, and Fig. \ref{fig:burgers:nf:ttf} shows the time-to-failure metric.
We see that the constrained approach has similar performance at just $\Nf = 500$ collocation points as at $\Nf = 1000$ or $\Nf = 2000$.
Comparing the left subplot of Fig. \ref{fig:burgers:nf:l2}, recall that \cite{raissi_deep_2018} originally uses $\Nf = 10000$, giving $\rell = 0.46$, at a quarter of the noise level ($0.05$, as compared to $0.20$).
Of course, as argued in Sec. \ref{sec:training:importance_of_simult}, the simultaneous training, which \cite{raissi_deep_2018} even briefly mentioned, likely accounts for a substantial portion of this difference.
For the testing initial condition, the constrained method may appear to perform worse with $\Nf = 2000$ compared to $\Nf = 1000$, although we attribute this difference to having five samples.
In Figs. \ref{fig:burgers:nf:l2} and \ref{fig:burgers:nf:ttf}, there are several outliers.
From our experience, performing validation over more seeds helps to remove these outliers, although at a greater computational cost.
A less expensive method to avoid these outliers may be the focus of future work.

\subsection{Korteweg-De Vreis}
We consider another system, governed by the Korteweg-De Vreis (KdV) equation:
\begin{equation}
    \begin{aligned}
        &u_t = -u u_x - u_{xxx}, \quad  &-20 &< x < 20, \   0 < t < 40, \\
        &u(x,0) = u_0(x), \quad &-20 &\leq x \leq 20 \\
        &u(-20,t) = u(20,t), \quad &0 &\leq t \leq 40, \\
        &u_x(-20,t) = u_x(20,t), \quad &0 &\leq t \leq 40 .
    \end{aligned}
    \label{eq:kdv}
\end{equation}
Following \cite{raissi_deep_2018}, we use different versions of $u_0 (x)$ for training and testing, which are given as
\begin{equation}
    \begin{aligned}
        &\utrain (x) = - \sin \left ( \frac{\pi x}{20} \right ),\\
        &\utest (x) = \cos \left ( \frac{\pi x}{20} \right ) .
    \end{aligned}
    \label{eq:kdv:ics}
\end{equation}

The details are very similar to the example with Burgers' equation.
Again, we use $N_u = 10000$ data points for training/validation, where these points are randomly sampled from a mesh of $n_x = 256$ points with $n_t = 200$ time steps.
The first two-thirds of the data are used for training, and the last third is used for validation to select from $10$ of the $\varepsilon$ of $\lambda$ hyperparameter and then from $3$ seed initializations, as given in Alg. \ref{alg:train_validation}.
We use the same architectures for $\Np$ and $\ut$ as above, with 2 hidden layers of 16 units and 5 hidden layers of 32 units respectively.
Now, we include a third spatial derivative in the potential PDE, $\Np (u, u_x, u_{xx}, u_{xxx})$.
As we have pointed out, and as others have studied \cite{raissi_deep_2018}, how many inputs and which inputs to include in $\Np$ requires careful considerations in practice.
We omit that discussion here.
We compare the two training methods, again over ensembles, varying the noise level and the number of collocation points, $\Nf$.

\subsubsection{Noise Level Study}
\begin{figure}[htb!]
    \centering
    \includegraphics[width=1.0\linewidth]{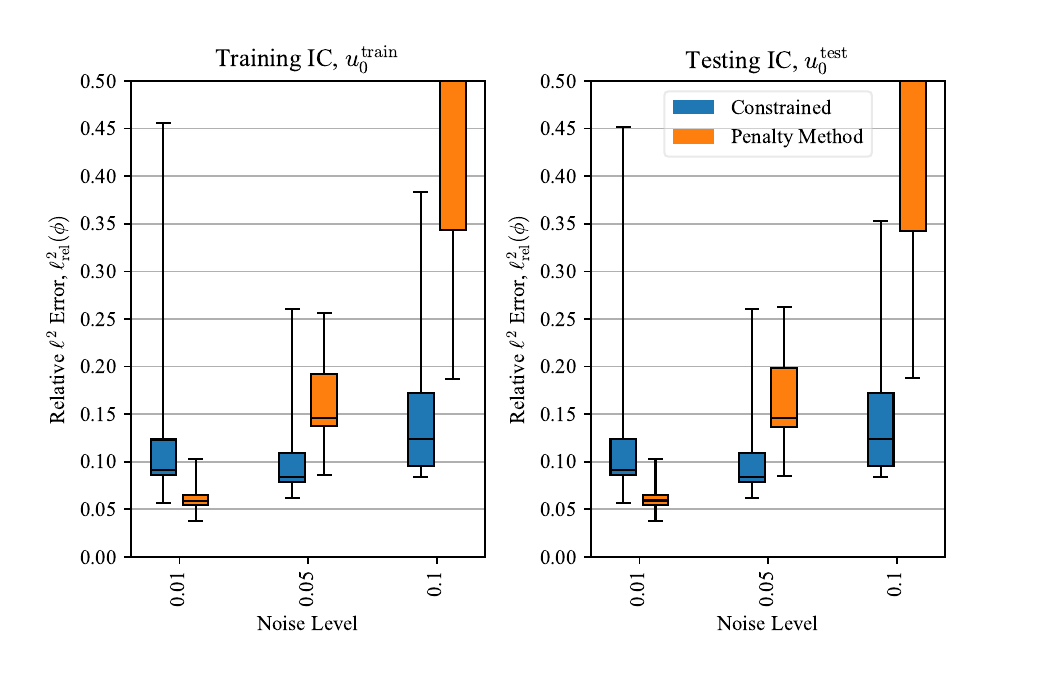}
    \caption{For the underlying KdV system, with $\Nf = 1000$ collocation points, the proposed constrained equation-discovery formulation outperforms the penalty-method equation-discovery approach, for larger noise levels. Each box-and-whisker plot shows 10 runs, like in Fig. \ref{fig:burgers:noise_level:l2}.}
    \label{fig:kdv:noise_level:l2}
\end{figure}

\begin{figure}[htb!]
    \centering
    \includegraphics[width=1.0\linewidth]{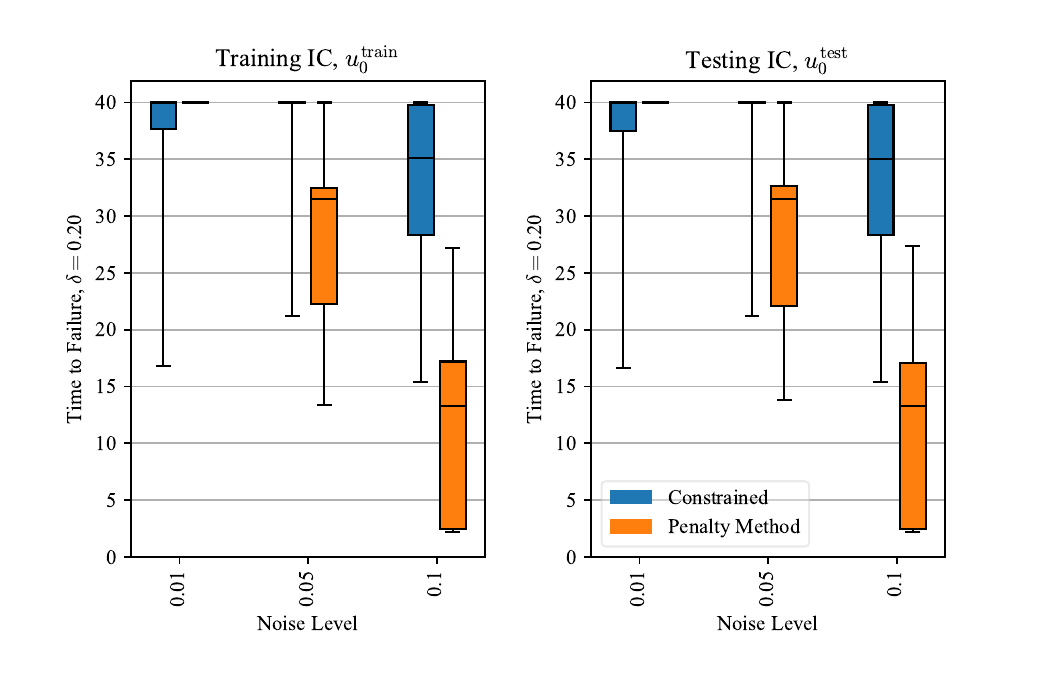}
    \caption{With the same setup and runs as Fig. \ref{fig:kdv:noise_level:l2}, we show the time-to-failure (Eq. (\ref{eq:results:ttf})). Higher is better for this performance measure. The maximum simulation is $T = 40$ for both $\utrain$ and $\utest$.}
    \label{fig:kdv:noise_level:ttf}
\end{figure}

We repeat the study presented for the Burgers' system in Sec. \ref{sec:burg:nl}, now for the KdV system of Eq. (\ref{eq:kdv}) with the training and testing initial conditions given by Eq. (\ref{eq:kdv:ics}).
As above, we repeat Alg. \ref{alg:train_validation} ten times for each box-and-whisker in Figs. \ref{fig:kdv:noise_level:l2} and \ref{fig:kdv:noise_level:ttf}.

For this problem, we solve the learned PDEs on an equispaced grid of $n_x = 64$ points, with $dt = 0.01 \cdot dx$, and a 4th-order Runge-Kutta time integrator, to generate the errors in Fig. \ref{fig:kdv:noise_level:l2}.
The validation meshes referred to in Eq. (\ref{eq:validation:final}) in Sec. \ref{sec:solving_nn_pdes} arbitrarily use $n_{x}^{1} = 56$, $n_{x}^{2} = 64$, and $n_{x}^{3} = 72$ grid points, again each adjusting $dt$ to follow the same $0.01$ ratio.
For the spatial derivatives, with periodic boundary conditions, we use 9-point finite difference schemes.
For the first two spatial derivatives, these schemes are both 8th order, but for the third derivative, the scheme is 6th order.
This scheme requires assuming higher-order periodicity than what Eq. (\ref{eq:kdv}) explicitly includes.

In Fig. \ref{fig:kdv:noise_level:l2}, we see a similar trend as in Fig. \ref{fig:burgers:noise_level:l2}, for the Burgers' example.
Here, the constrained method is more robust to noise than the penalty method.
Note that unlike the Burgers' example, at the lowest noise level, the penalty method outperforms the constrained approach.
Here, we also see better generalization from $\utrain$ to $\utest$, due to the similarity of the initial conditions in Eq. (\ref{eq:kdv:ics}).
Fig. \ref{fig:kdv:noise_level:ttf} shows the time-to-failures for the same runs, again supporting the strength of the constrained method for higher noise levels.

\begin{figure}[htb!]
    \centering
    \includegraphics[width=1.0\linewidth]{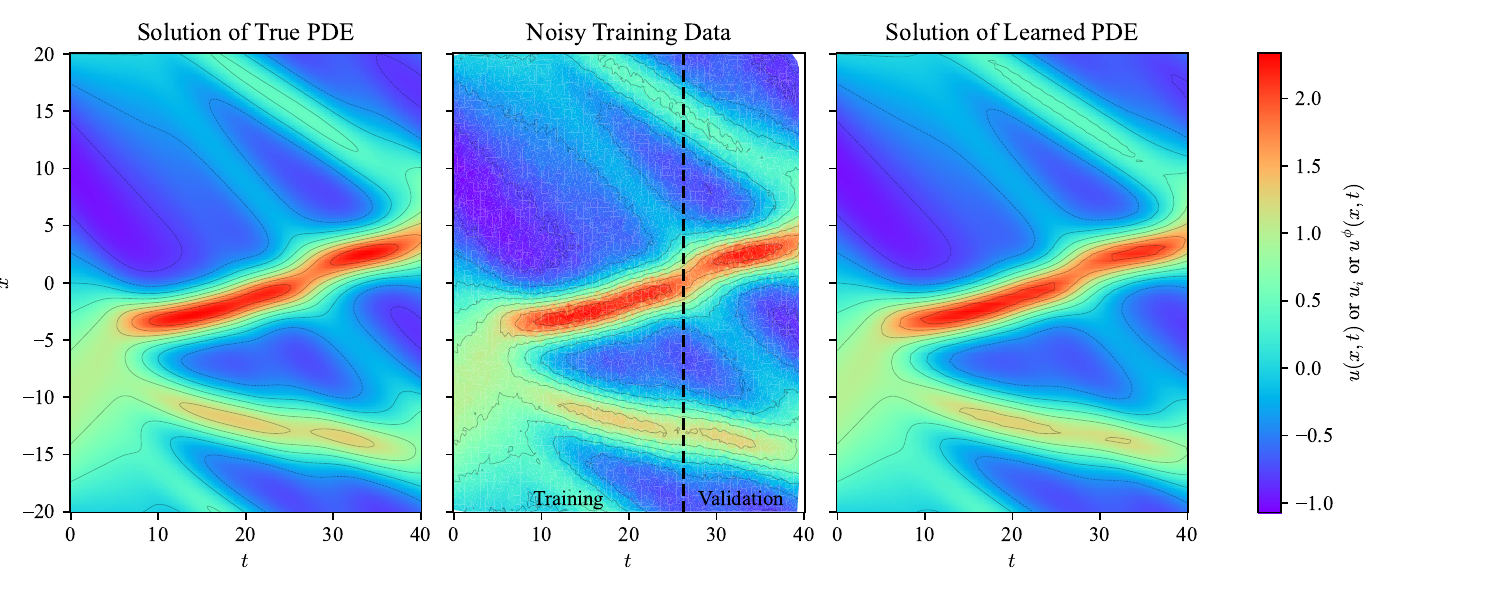}
    \caption{Here, we compare the solution of the true PDE (left), to the solution of a discovered PDE (right). The PDE was discovered by applying the constrained method to the training data (middle). We have arbitrarily chosen to show the lowest training error from the $0.05$ noise level in Fig. \ref{fig:kdv:noise_level:l2}, for the constrained formulation.}
    \label{fig:kdv:example:con:train}
\end{figure}

\begin{figure}[htb!]
    \centering
    \includegraphics[width=1.0\linewidth]{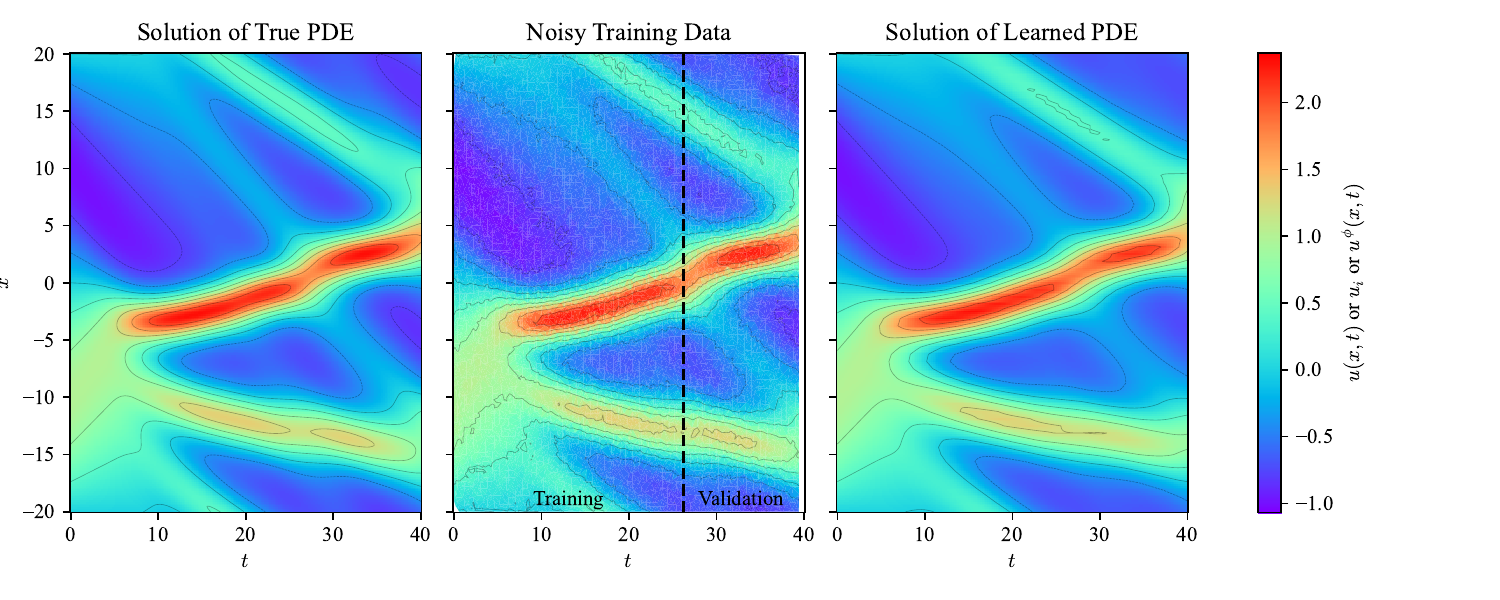}
    \caption{We show the same case as Fig. \ref{fig:kdv:example:con:train}, but where the PDE was discovered through the penalty-method formulation. This is the lowest training error from the $0.05$ noise level in Fig. \ref{fig:kdv:noise_level:l2}, for the penalty-method formulation.}
    \label{fig:kdv:example:sa:train}
\end{figure}

Figs. \ref{fig:kdv:example:con:train} and \ref{fig:kdv:example:sa:train} show the true solution, the noisy training data, and the solution to a discovered PDE.
Fig. \ref{fig:kdv:example:con:train} shows the case from the lowest training error for the constrained method at a noise level of $0.05$.
Fig. \ref{fig:kdv:example:sa:train} shows the result from the lowest training error, but for the penalty-method approach.
Due to the comparable performance between $\utrain$ and $\utest$, we do omit showing the solution of these PDEs on $\utest$.
Both cases show discovered PDEs that admit solutions comparable to the solution of the true PDE.

\subsubsection{Number of Collocation Points, $\Nf$}
\begin{figure}[htb!]
    \centering
    \includegraphics[width=0.85\linewidth]{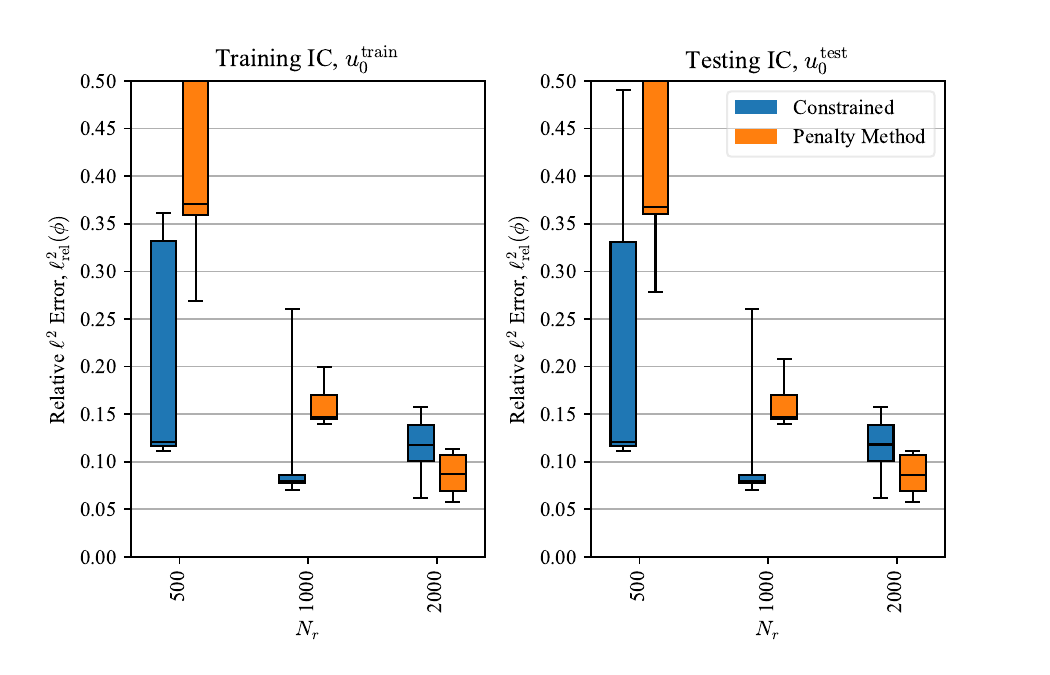}
    \caption{For the underlying KdV system, with a fixed noise level of $0.05$, we compare the performance of the two methods over changes to the number of collocation points, $\Nf$. Each box-and-whisker plot includes 5 runs, so each run is shown by a different feature.}
    \label{fig:kdv:nf:l2}
\end{figure}

We repeat the above, but now holding the noise level at the middle level of $0.05$, and varying the number of collocation points, $\Nf$. Everything else is exactly the same.
As opposed to Fig. \ref{fig:kdv:noise_level:l2}, the results in Fig. \ref{fig:kdv:nf:l2} include five, rather than ten repetitions of Alg. \ref{alg:train_validation} (for each box-and-whisker).

\begin{figure}[htb!]
    \centering
    \includegraphics[width=0.85\linewidth]{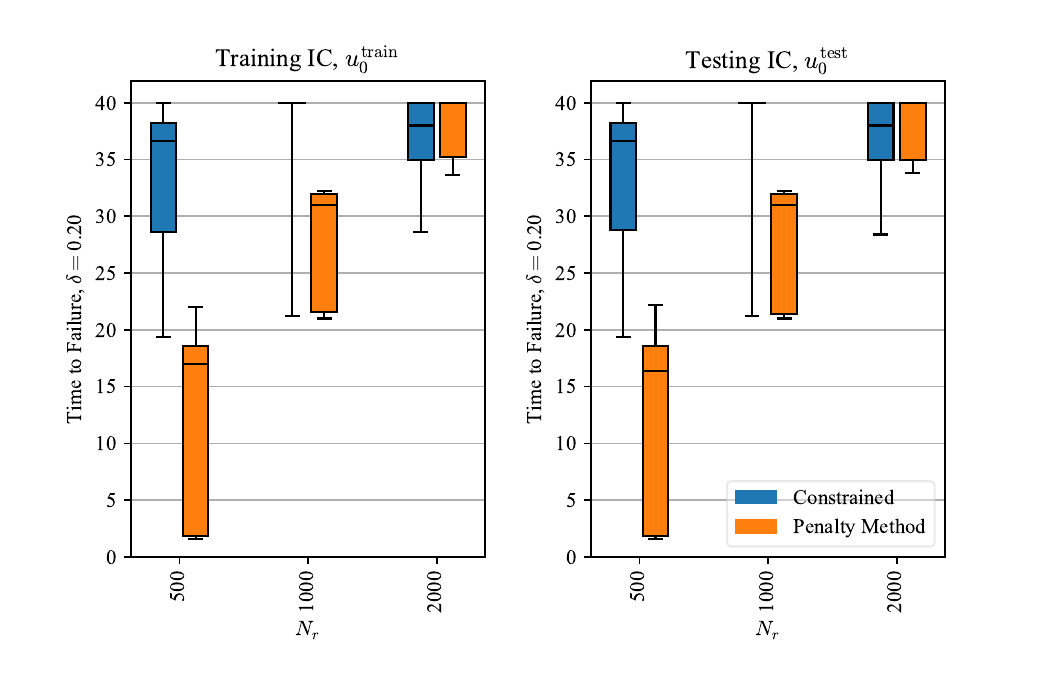}
    \caption{With the same setup and runs as Fig. \ref{fig:kdv:nf:l2}, we show the time-to-failure (Eq. (\ref{eq:results:ttf})). Higher is better for this performance measure. The maximum simulation is $T = 40$ for both $\utrain$ and $\utest$.}
    \label{fig:kdv:nf:ttf}
\end{figure}

Fig. \ref{fig:kdv:nf:l2} shows the quick improvement of the penalty method, when more collocation points are added.
For $\Nf = 500$, the constrained approach performs better than the penalty method, but at $\Nf = 2000$, the penalty method instead has lower errors than the constrained approach.
In Fig. \ref{fig:kdv:nf:ttf}, with $\Nf = 2000$ collocation points, three of the five instances of the penalty approach do not exceed the threshold $\delta$ for any time, for both the training and testing initial conditions.

\section{Discussion and Conclusion}
In this section, we briefly summarize and extend the main points and contributions of this work.

\subsection{Constrained Approach vs. Penalty-Method Approach}
In this work, we presented two training methods for finding a neural network $\Np$ describing the time evolution of some system as a PDE, using an intermediate neural network $\ut$ to represent the state.
We generalized Self-Adaptive PINNs to this PDE discovery framework \cite{Mcclenny2022}, and proposed a new method, which solves the same constrained optimization problem (Eq. (\ref{eq:constrained:constrained_form})) through a widely used trust-region barrier method (See \ref{sec:sqp} \cite{byrd_interior_1999}), still only with first-order derivative information through BFGS approximations of the Hessians \cite{nocedal_numerical_2006}.

\subsubsection{Performance}
\label{sec:discussion:perf}

In Fig. \ref{fig:burgers:noise_level:l2}-\ref{fig:kdv:nf:ttf}, we compare the constrained approach to the penalty method and see that for high noise levels and few collocation points, the constrained method outperforms the penalty method.
This performance is evaluated through numerically solving the discovered PDE with the method of lines, for both the training initial condition and an unseen testing or extrapolation initial condition.
We looked at standard relative $\ell^2$ errors and a time-to-failure metrics, which supported one another.

For the Burgers' equation example, both methods appear to be more robust to noise than the $\rell = 0.46$ reported in \cite{raissi_deep_2018} for a similar problem for $\text{noise level } = 0.05$, while using $10 \times$ fewer collocation points (compare Fig. \ref{fig:burgers:noise_level:l2} to Table 1 in \cite{raissi_deep_2018}).
However, the penalty-method approach is very similar to the method presented therein, just fixing $\lambda = 1$ for all collocation points.
Thus, we attribute this improved robustness to simultaneously training $\ut$ and $\Np$, which was briefly suggested in \cite{raissi_deep_2018}, although not explored.
In Sec. \ref{sec:training:importance_of_simult} we argued for simultaneous training from a theoretical, regularization perspective, but in Sec. \ref{sec:burgers} we observed strong performance in the presence of noise, unlike in \cite{raissi_deep_2018}.

\subsubsection{Computational Cost}
However, the two presented methods differ in computational cost.
As implemented, the penalty method is substantially faster and requires less memory.
The constrained method constructs the Jacobian matrix for its constraints and builds BFGS-approximations to their (nonsparse) Hessian matrices as well (or really their inverses).
Conversely, the penalty-method approach computes the same gradients of the constraints, but accumulates them just into the gradient of the entire loss function, updating the parameters through fewer, simpler operations.
In other words, both methods use gradients of the $\fj$ terms, but the penalty method immediately agglomerates this information over $j \in \{ 1, 2, ..., \Nf \}$, and then combines it with the data-matching mean squared error.
The constrained method eventually agglomerates this information into a single update for $\theta$ and $\phi$, but it first uses these large matrices in the trust-region, barrier method, involving several linear systems that scale based on the number of constraints.
Thus, the constrained method depends more heavily on the number of collocation points $\Nf$ than the penalty method, although the constrained method works well with fewer collocation points.
The linear solves increase both the memory requirements and the run time of the constrained method's iterations, but both methods are still fundamentally first-order, only using gradient information of the constraints and the objective.

Currently, the constrained method is substantially more computationally expensive, because of both the method itself and its software implementation.
There is a considerable slowdown because of the communication between different scientific computing and deep learning frameworks (SciPy / NumPy vs. PyTorch).
However, given the massive growth of the area in recent years \cite{cuomo_scientific_2022, montans_data-driven_2019}, and the current work specifically in the area of classical solvers for deep learning frameworks \cite{zampini_petscml_2024}, this downside may soon be mitigated.
Further, other improvements in PINNs may generally be applied to this method too, for instance with domain decomposition for parallelization \cite{shukla_parallel_2021} or by enforcing initial and boundary conditions strongly in the design of $\ut$ \cite{lu_physics-informed_2021}.

We also claim that, in practice, the collection of the training data $\{ (x_i, t_i), u_i \}_{i=1}^{N_u}$ may outweigh the cost of the equation discovery.
For instance, if this data comes from some experiment, performing the experiment itself is likely substantially more expensive than this equation-discovery (post-processing) step.
In this scenario, ensuring that the equation-discovery method is robust to noise will be more important than having a shorter runtime.
This situation also highlights when this method may be preferred over operator or flow map learning \cite{lu_deeponet_2021, li_fourier_2021, churchill_flow_2023, wu_data-driven_2020, chen_deep_2022}.
As mentioned in Sec. \ref{sec:intro}, these methods learn both the PDE and its solution (as one object).
If obtaining the experimental data is expensive, generating enough data for these methods quickly becomes infeasible.

\subsection{Simultaneous Training}
As motivated in Sec. \ref{sec:training:importance_of_simult} and empirically supported in Sec. \ref{sec:burg:nl}, simultaneous training of $\ut$ and $\Np$ could improve robustness to the noise in the data.
Despite the neural network $\ut$ being smoother than the original noisy data, it is still subject to some overfitting.
Further, even automatic differentiation of $\ut$ worsens this issue.
While other methods exist to regularize the training of $\ut$ or its differentiation \cite{goodfellow_deep_2016, Chartrand2011, cortiella_priori_2022}, we have shown reasonable success using $\Np$ in a Sobolev-like regularization term during simultaneous training of $\ut$ and $\Np$.

\subsection{Validation Findings}
In both the constrained and penalty formulations, a fundamental hyperparameter emerges, weighting the importance of the data against the learned PDE.
For our numerical examples, the random initialization also plays an important role in the performance.
To select these hyperparameters, we follow the machine learning convention with validation.
As discussed in Sec. \ref{sec:validation}, there are multiple notions of validation that could be used.
We propose to split the data into training and validation splits, such that the validation set comes directly after the training set, with respect to the simulation time $t$ \cite{cerqueira_evaluating_2020}.
We also suggest comparing the numerical solution of the PDE to the data as in Eq. (\ref{eq:solving_nn_pdes}) and (\ref{eq:validation}).
This validation approach is consistent with our use of $\Np$ with classical solvers.

Further Fig. \ref{fig:refinements:burgers:noise_level:l2}-\ref{fig:refinements:kdv:nf:l2} in \ref{sec:refinement_figures} show the robustness of the learned PDEs to changes in the numerical solver (specifically just $n_x$).
This shows that the temporary validation loss over multiple meshes (Eq. (\ref{eq:validation:final})) was at least partially successful.

\subsection{Classical Solution Methods}
We elect to use a classical numerical solver for $\Np$, as opposed to a PINNs solver like in \cite{raissi_deep_2018}.
Even on well-studied PDEs, PINNs solvers are still poorly understood, unpredictable, and comparatively expensive \cite{chuang_experience_2022, cuomo_scientific_2022}, partially motivating our choice.
Also, this work demonstrates that these neural network PDEs can be solved through classical methods, although a number of new challenges emerge.
Foremost, the solution is often quite sensitive to the spatial discretization (yet in our experience, comparatively robust to the temporal discretization).
For simplicity, we use a temporary fix of computing the validation loss from multiple meshes to promote robustness (Eq. (\ref{eq:validation:final})).
While right now this is a major issue, it may be the focus of future work.
Others have already demonstrated some success in this area, by learning the numerical scheme in addition to the PDE \cite{long_pde-net_2019}, or by learning stable numerical schemes \cite{prakash_data-driven_2024}.
We may also investigate generalizing this approach to a probabilistic setting, attempting to capture uncertainty in the governing PDE.

\section*{CRediT authorship contribution statement}
\textbf{Grant Norman:} Conceptualization, Data curation, Formal analysis, Investigation, Methodology, Software, Validation, Visualization, Writing -- original draft, Writing -- review \& editing.
\textbf{Jacqueline Wentz:} Conceptualization, Formal analysis, Methodology, Writing -- review \& editing.
\textbf{Hemanth Kolla:} Conceptualization, Formal analysis, Writing -- review \& editing.
\textbf{Kurt Maute:} Conceptualization, Formal analysis, Writing -- review \& editing.
\textbf{Alireza Doostan:} Conceptualization, Formal analysis, Funding acquisition, Investigation, Methodology, Project administration, Resources, Validation, Writing -- review \& editing.

\section*{Declaration of competing interest}

Grant Norman reports financial support was provided by Department of Energy. Alireza Doostan reports financial support was provided by Department of Energy. If there are other authors, they declare that they have no known competing financial interests or personal relationships that could have appeared to influence the work reported in this paper.

\section*{Acknowledgements}
The authors would like to thank Juan M. Cardenas (CU Boulder) and Reese E. Jones (Sandia National Laboratories) for their helpful comments and insights related to this work.

This work was supported by the Department of Energy, National Nuclear Security Administration, Predictive Science Academic Alliance Program (PSAAP), Award Number DE-NA0003962.
Sandia National Laboratories is a multi-mission laboratory managed and operated by National Technology \& Engineering Solutions of Sandia, LLC (NTESS), a wholly owned subsidiary of Honeywell International Inc., for the U.S. Department of Energy’s National Nuclear Security Administration (DOE/NNSA) under contract DE-NA0003525. This written work is authored by an employee of NTESS. The employee, not NTESS, owns the right, title and interest in and to the written work and is responsible for its contents. Any subjective views or opinions that might be expressed in the written work do not necessarily represent the views of the U.S. Government. The publisher acknowledges that the U.S. Government retains a non-exclusive, paid-up, irrevocable, world-wide license to publish or reproduce the published form of this written work or allow others to do so, for U.S. Government purposes. The DOE will provide public access to results of federally sponsored research in accordance with the DOE Public Access Plan.


\clearpage
\appendix
\section{}

\subsection{Neural Networks}
\label{sec:background:nns}

A neural network is an arbitrary function approximator as a sequence of linear and nonlinear compositions \cite{goodfellow_deep_2016}, taking the general form
\begin{equation}
    \Phi (z) = W_n ( \phi_{n-1} \circ \phi_{n-2} \circ \cdots \circ \phi_0 (z) ) + b_n, \quad z \to \phi_i(z) = \sigma (W_i z + b_i),
\end{equation}
where we will use the activation function of $\sigma (\cdot) = \sin (\cdot)$, as presented in \cite{sitzmann_implicit_2020}.
The set of weights and biases, $\left \{ W_n, b_n, \hdots, W_0, b_0 \right \}$, are the parameters of the neural network and are updated through gradient-based optimization.
The gradients of these parameters are calculated through backpropagation (reverse-mode automatic differentiation).
Using this same technique, we can compute the gradient of the neural network's output with respect to the inputs, $\Phi_z$.

\subsection{Trust-Region Sequential Quadratic Program}
\label{sec:sqp}
As mentioned in Sec. \ref{sec:constrained_method}, the original inequality constrained optimization problem from Eq. (\ref{eq:constrained:loose_constrained_form}) is converted to a sequence of equality-constrained optimization problems by a barrier method, with the introduction of slack variables.
Each equality-constrained problem is solved through a trust-region Sequential Quadratic Program (SQP) \cite{byrd_interior_1999}.
The SQP from \cite{byrd_interior_1999} is repeated here, and its implementation is available through SciPy \cite{virtanen_scipy_2020, lalee_implementation_1998}.
We additionally approximate the Hessians through BFGS updates \cite{nocedal_numerical_2006}.

For simplicity, we will adjust our notation for the context of constrained optimization.
We group the optimization parameters $\theta, \phi$ as $x$, and give our objective function as $h(x)$, and our inequality constraints as the vector-valued $g(x)$.
In particular, from Eq. (\ref{eq:constrained:loose_constrained_form}),
\begin{align*}
    h(x) & = h \left ( \left [ \theta, \phi \right ]^T \right ) = \frac{1}{N_u} \sum_{i=1}^{N_u} \left (  u_i - u^\theta (x_i,t_i) \right )^2, \\
    g_j(x) & = g_j \left ( \left [ \theta, \phi \right ]^T \right ) = 
             \left ( \ut_t(x_j,t_j) - \left . \Np(\ut, \ut_x, \ut_{xx}) \right | _{(x_j,t_j)} \right ) -\varepsilon, \quad j = 1,2,\hdots,\Nf, \\
    g_{j+\Nf}(x) & = g_{j+\Nf} \left ( \left [ \theta, \phi \right ]^T \right ) = 
             - \left ( \ut_t(x_j,t_j) - \left . \Np(\ut, \ut_x, \ut_{xx}) \right | _{(x_j,t_j)} \right ) -\varepsilon, \quad j = 1,2,\hdots,\Nf, \\
    \label{eq:sqp:substitutions}
\end{align*}
The constrained optimization problem from Eq. (\ref{eq:constrained:loose_constrained_form}) can be restated as
\begin{equation}
    \begin{aligned}
        & \min_{x} \quad h(x), \\
        & \text{subject to} \quad g_j (x) \leq 0, \quad j = 1,2,\hdots,2 \Nf.
    \end{aligned}
    \label{eq:sqp:loose_constrained_form}
\end{equation}
By introducing slack variables $s$ with a barrier method and penalty parameter $\mu > 0$, Eq. (\ref{eq:sqp:loose_constrained_form}) is converted into an equality-constrained problem,
\begin{equation}
    \begin{aligned}
        & \min_{x} \quad h(x) - \mu \sum_{j=1}^{2 \Nf} \log s_i, \\
        & \text{subject to} \quad g_j (x) + s_j = 0, \quad j = 1,2,\hdots,2 \Nf.
    \end{aligned}
    \label{eq:sqp:eq_constrained_form}
\end{equation}
The Lagrangian of Eq. (\ref{eq:sqp:eq_constrained_form}), with multipliers $\nu$, is given as
\begin{equation}
    L (x,s,\nu) = h(x) - \mu \sum_{j=1}^{2 \Nf} \log s_j + \nu^T (g(x) + s).
    \label{eq:sqp:lagrangian}
\end{equation}
At each iteration of the SQP, a quadratic approximation of the Lagrangian (Eq. (\ref{eq:sqp:lagrangian})) is constructed, and a step is proposed to minimize this approximation, also subject to a linearization of the constraints and an iteratively updated bound (the trust region).
If the proposed step sufficiently decreases the optimimality conditions, the step is accepted.
These conditions are specifically the KKT conditions,
\begin{equation}
    \begin{aligned}
        \nabla_x L (x,s,\nu)&  = E_1, \\
        s_j \cdot  \nabla_{s_j} L (x,s,\nu) & = E_2, \quad j=1,2,...,2 \Nf, \\
        g(x) + s & = E_3.
    \end{aligned}
    \label{eq:sqp:kkts}
\end{equation}
The second set of conditions arise by multiplying the gradient of the Lagrangian with respect to $s$ by $s$.
If $E_2 = 0$, then both the original condition and this modified condition are satisfied.
This is referred to as the ``perturbed KKT system'' in \cite{byrd_interior_1999}.
These conditions also form the basis for the SQP step.
The proposed step is effectively a Newton step towards the solution of this system, with $E_1 = E_2 = E_3 = 0$.
This step is taken over $x$, $s$, and $\nu$.
Such a step requires gradients of the left-hand sides of Eq. (\ref{eq:sqp:kkts}), thus requiring $\nabla_{xx} L$ and $\nabla_{s s} L$.

From Eq. (\ref{eq:sqp:lagrangian}), we can compute $\nabla_{s s} L = \mu \cdot \mathrm{ diag }(s) ^{-2}$.
An approximation of $\nabla_{s s} L$ is used, by again assuming $\nabla_{s_j} L = \nu_j - \mu s_j^{-1} = 0$.
Solving and substituting $\mu = \nu_j s_j$ for each diagonal entry in $\mu \cdot \mathrm{diag} (s)^{-2}$ gives $\nabla_{s s} L = \mathrm{diag} (s) ^{-1} \mathrm{diag} (\nu)$.
The Lagrange multipliers are estimated through a least squares estimate.
This could lead to nonpositive Lagrange multipliers.
In the case where a given Lagrange multiplier, $\nu_j$, is nonpositive, that entry of $\nabla_{s s} L$ is replaced with the original $\mu \cdot s_{j}^{-2}$ from above.
Next, the computation of $\nabla_{xx} L$ subsequently requires $\nabla_{xx} h (x)$ and $\nabla_{xx} g_j (x)$ for $j = 1,2,..., 2 \Nf$.
We use the BFGS iterative update to approximate these Hessians.

The SQP described above proposes a change to $x$ and to $s$, denoted $d_x$ and $d_s$ respectively.
If these changes fall within the trust region, then they are accepted and the optimization and slack variables are updated.
The goal is to restrict changes to the domain where the above quadratic model of the Lagrangian is a good approximate of the underlying Lagrangian.
The step $d_x$ and the scaled $\mathrm{diag} (s)^{-1} d_s$ are together restricted in the Euclidean norm by some trust region radius.
Further, the \textit{fraction to the boundary} rule is used as a second constraint to maintain $s > 0$.

Now, to solve the quadratic program, it is broken up into two parts: normal and tangential (or vertical and horizontal).
The normal step aims to satisfy the linear constraints ($g(x) + s = 0$), and the horizontal step is taken tangent to the constraint space, aiming for optimality.
The normal or vertical step is computed with a modified dogleg procedure \cite{nocedal_numerical_2006}, which constructs and solves a linear system without iteration.
The tangential or horizontal step is then computed, and its solution is the most expensive part of the whole method.
After constructing a modified horizontal subproblem, a projected conjugate gradient method is used to solve it iteratively.

Following the solution of an iteration of the SQP, a merit function (which is a slight modification of $L (x,s,\nu)$) is evaluated and used to compute the update to update its own penalty parameter, through a predicted reduction.
The predicted reduction is compared to the actual reduction, which is used to update the trust-region radius, or to continue another iteration.
Further, based on these two reductions and if the normal step is small compared to the tangential step, then a second-order correction term is added, which requires the solution of more linear systems.

For further details beyond this brief overview, the reader is directed to \cite{byrd_interior_1999, lalee_implementation_1998, nocedal_numerical_2006, virtanen_scipy_2020}.

\clearpage
\section{Numerical Solutions with Different Meshes}
\label{sec:refinement_figures}
This section shows the same models for $\Np$ as in Fig. \ref{fig:burgers:noise_level:l2}, \ref{fig:burgers:nf:l2}, \ref{fig:kdv:noise_level:l2}, and \ref{fig:kdv:nf:l2}.
However, instead of showing the results for a single numerical solve with a fixed number of grid points, we vary $n_x$ (and consequently also $n_t$.
The purpose of this is to examine how sensitive $\Np$ is to the numerical solution methodology.
If the figures are the same across each row, that means that the numerical solution does not substantially change under the given mesh refinements.
Note, the mesh refinements selected for this study are different than the validation meshes ($n_{x}^{1}$, $n_{x}^{2}$, $n_{x}^{3}$) used to promote robustness (See Sec. \ref{sec:solving_nn_pdes} and Eq. (\ref{eq:validation:final})).
In all of these, $n_t$ is taken so that the ratio between the mesh size and the time step remains the same.
This ratio is the same ratio used in Sec. \ref{sec:solving_nn_pdes}.
See the code implementation on GitHub (upon acceptance) for further details.

Generally, we see that the instances that are sensitive to $n_x$ begin with higher $\rell$ errors.
On the other hand, results with low $\rell$ tend to be robust to changes in $n_x$.
This gives hope that by decreasing $\rell$ for a given mesh, the discovered $\Np$ will generalize well to other meshes.

\subsection{Burgers' Equation}
\begin{figure}[htb!]
    \centering
    \includegraphics[width=0.87\linewidth]{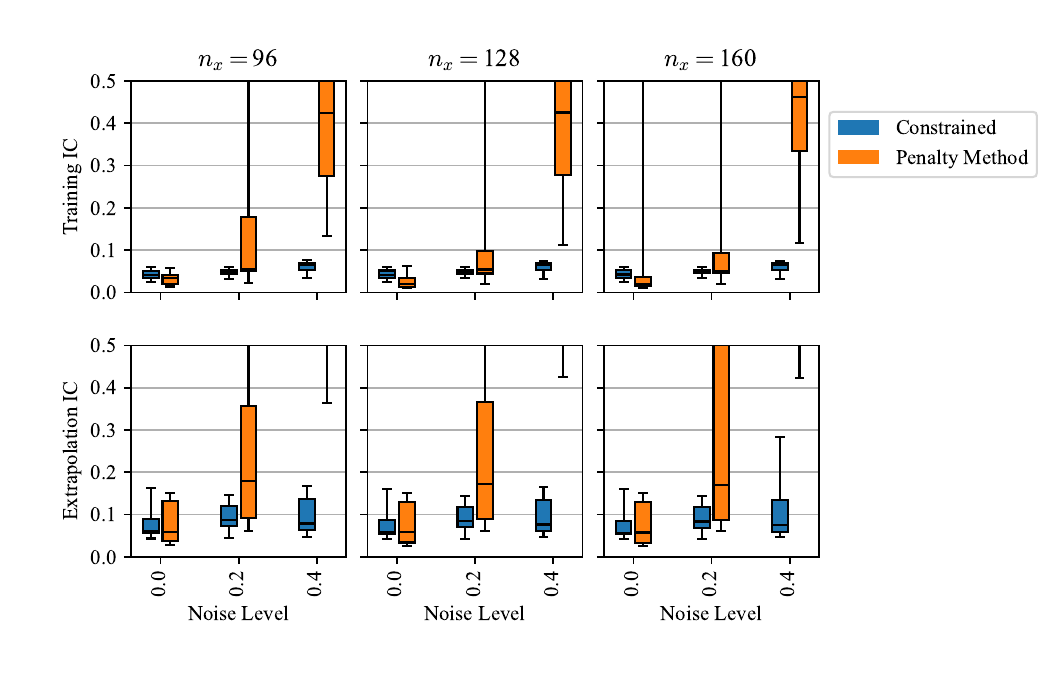}
    \caption{The results from Fig. \ref{fig:burgers:noise_level:l2} are shown in the middle column. The y-axis shows the same error metric, $\rell$.}
    \label{fig:refinements:burgers:noise_level:l2}
\end{figure}

\begin{figure}[htb!]
    \centering
    \includegraphics[width=0.87\linewidth]{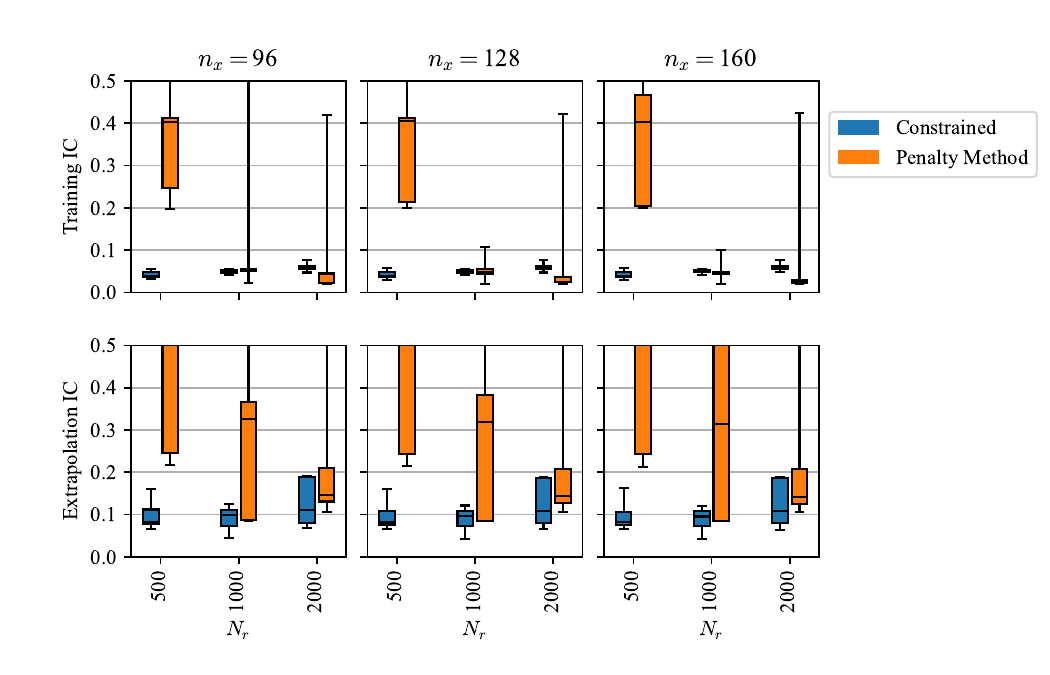}
    \caption{The results from Fig. \ref{fig:burgers:nf:l2} are shown in the middle column. The y-axis shows the same error metric, $\rell$.}
    \label{fig:refinements:burgers:nf:l2}
\end{figure}

\clearpage
\subsection{KdV Equation}
\begin{figure}[htb!]
    \centering
    \includegraphics[width=0.87\linewidth]{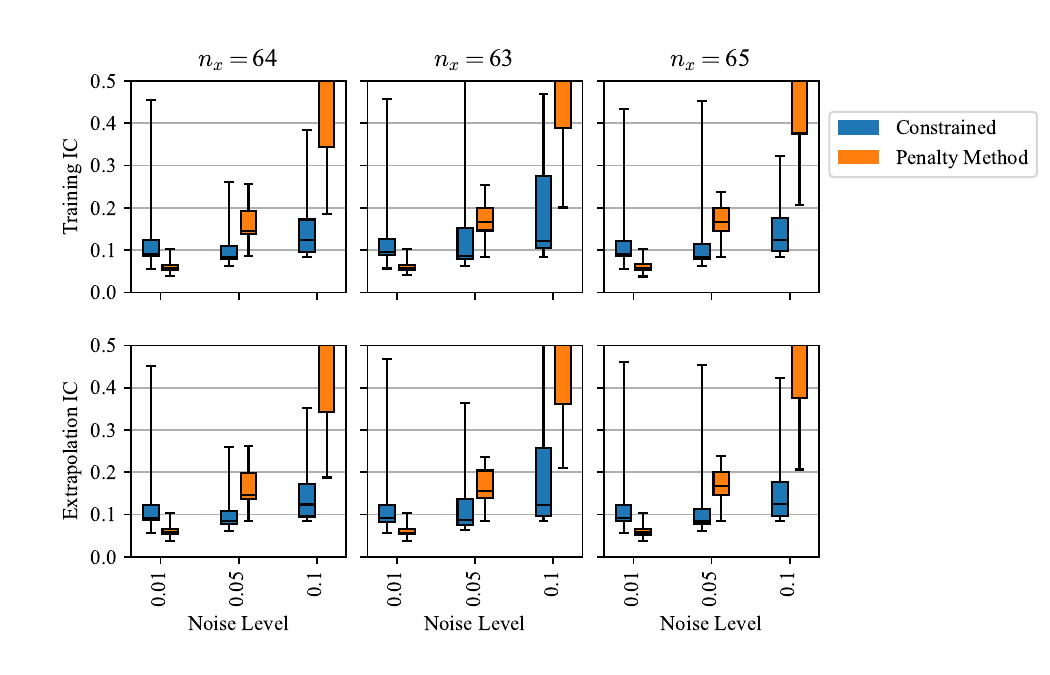}
    \caption{The results from Fig. \ref{fig:kdv:noise_level:l2} are shown in the left column. The y-axis shows the same error metric, $\rell$.}
    \label{fig:refinements:kdv:noise_level:l2}
\end{figure}

\begin{figure}[htb!]
    \centering
    \includegraphics[width=0.87\linewidth]{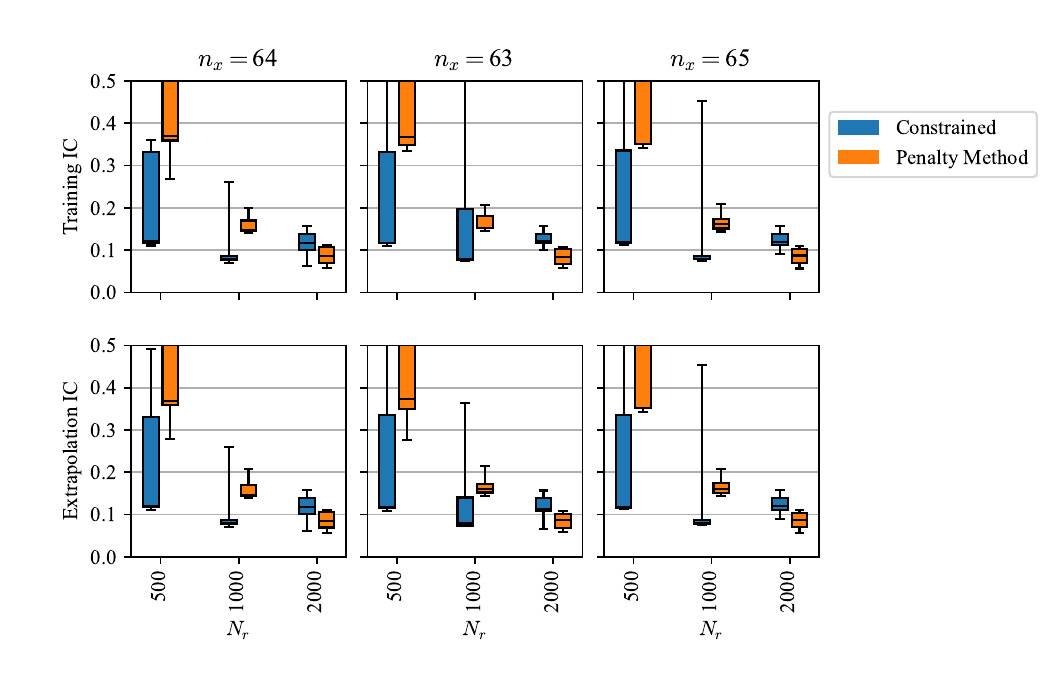}
    \caption{The results from Fig. \ref{fig:kdv:nf:l2} are shown in the left column. The y-axis shows the same error metric, $\rell$.}
    \label{fig:refinements:kdv:nf:l2}
\end{figure}



\clearpage
\bibliographystyle{elsarticle-num}
\bibliography{zotero_all}


\end{document}